\pdfoutput=1
\documentclass[11pt]{article}

\usepackage[final]{coling}

\usepackage{times}
\usepackage{latexsym}

\usepackage[T1]{fontenc}
\usepackage[utf8]{inputenc}

\usepackage{microtype}

\usepackage{inconsolata}

\usepackage{graphicx}
\usepackage{amsmath}
\usepackage{amssymb}
\usepackage{booktabs} 
\usepackage{algorithm}
\usepackage{algorithmic}

\title{Representation Purification for End-to-End Speech Translation}

\author{
 \textbf{Chengwei Zhang\textsuperscript{1,2,\dag}},
 \textbf{Yue Zhou\textsuperscript{1,2,\dag}},
 \textbf{Rui Zhao\textsuperscript{1,2,3}},
 \textbf{Yidong Chen\textsuperscript{1,2,3}},
 \textbf{Xiaodong Shi\textsuperscript{1,2,\thanks{Corresponding author.}}}
 \\
 \textsuperscript{1}School of Informatics, Xiamen University, China\\
 \textsuperscript{2}Key Laboratory of Digital Protection and Intelligent Processing of Intangible Cultural Heritage \\of Fujian and Taiwan, Ministry of Culture and Tourism, China \\
 \textsuperscript{3}Key Laboratory of Multimedia Trusted Perception and Efficient Computing, \\Ministry of Education of China, Xiamen University, Xiamen, China
 \\
 \texttt{\{cwzhang98, zhouyue1, zhsqzr\}@stu.xmu.edu.cn, \{ydchen, mandel\}@xmu.edu.cn}
}

\begin{document}
\maketitle
\def\thefootnote{\dag}\footnotetext{Equal contribution.}\def\thefootnote{\arabic{footnote}}
\begin{abstract}
Speech-to-text translation (ST) is a cross-modal task that involves converting spoken language into text in a different language.
Previous research primarily focused on enhancing speech translation by facilitating knowledge transfer from machine translation, exploring various methods to bridge the gap between speech and text modalities.
Despite substantial progress made, factors in speech that are not relevant to translation content, such as timbre and rhythm, often limit the efficiency of knowledge transfer.
In this paper, we conceptualize speech representation as a combination of content-agnostic and content-relevant factors.  
We examine the impact of content-agnostic factors on translation performance through preliminary experiments and observe a significant performance deterioration when content-agnostic perturbations are introduced to speech signals.
To address this issue, we propose a \textbf{S}peech \textbf{R}epresentation \textbf{P}urification with \textbf{S}upervision \textbf{E}nhancement (SRPSE) framework, which excludes the content-agnostic components within speech representations to mitigate their negative impact on ST.
Experiments on MuST-C and CoVoST-2 datasets demonstrate that SRPSE significantly improves translation performance across all translation directions in three settings and achieves preeminent performance under a \textit{transcript-free} setting.
\end{abstract}

\section{Introduction}
Speech-to-text translation (ST) task aims to translate source language speech into target language text.
Earlier conventional ST systems~\cite{sperber2017neural, sperber-etal-2019-self, indurthi-etal-2023-clad} typically cascade automatic speech recognition (ASR) and machine translation (MT) to perform ST, which may suffer from error propagation and high latency. Consequently, end-to-end (E2E) ST systems have gained increasing attention due to their potential to mitigate these deficiencies~\citep{wang2020fairseq, Wang_Wu_Liu_Yang_Zhou_2020, liu2020bridging, xu2021stacked, du2022regularizing}.

As a cross-modal task, ST encounters additional challenges compared to MT, as speech encompasses not only the content information necessary for translation but also other factors such as timbre and pitch.
Therefore, MT is often considered as the performance upper-bound of ST, prompting researchers to devote considerable effort to designing sophisticated methods for facilitating knowledge transfer from MT to ST, such as multi-task learning~\citep{ye21_interspeech, zhang2023rethinking, zhou2024multitask}, knowledge distillation~\citep{liu2019end, 10096899, lei-etal-2023-ckdst}, and cross-modal alignment~\citep{fang2022stemm, ye2022cross, zhou2023cmot, 10447926, pmlr-v202-le23a}.
However, the inherent information divergence between speech and text continues to hinder the efficiency of knowledge transfer and the generalization capability~\citep{chan2022contentcontextfactorizedrepresentationsautomated, 10447494}.
Despite impressive improvements achieved in previous research, the impact of redundant speech factors on ST models is often overlooked.

% \citet{10447494} implicitly mitigate this issue from an optimization perspective, they proposed to align the speech and text representation space rather than sample pairs, bypassing the root of the problem.
% Despite further improvements in transfer efficiency, the correlation between various speech factors and ST performance still remains unclear.
% However, the inherent information divergence between speech and text, caused by fundamental characteristics of speech, continues to hinder the efficiency of knowledge transfer. \citet{10447494} implicitly mitigate this issue from an optimization perspective, bypassing the root of the problem.

In this paper, we view speech as an amalgamation of information, and following previous works~\citep{pmlr-v119-qian20a, 9747763}, we conceptualize it as a composite of four components: language content, timbre, pitch, and rhythm. We define the language content as \textbf{content-relevant} information, which refers to the textual information contained in speech signals. Consequently, the other three components are defined as \textbf{content-agnostic} information. 
We first conduct a preliminary study (see Section~\ref{sec:preliminary_analysis}) to investigate the correlation between the model's performance and the content-agnostic information. 
We observed that the ST model is susceptible to perturbations in the content-agnostic aspects of speech, with a significant performance gap between using original and perturbed speech as input. 
Moreover, the translation quality declines rapidly as content-agnostic information increases.
Based on these findings, we aim to purify the speech representation by explicitly filtering out the content-agnostic components.

To achieve this, we propose the \textbf{S}peech \textbf{R}epresentation \textbf{P}urification with \textbf{S}upervision \textbf{E}nhancement (\textbf{SRPSE}) framework.
Specifically, we introduce a content-agnostic encoder and a complex-information encoder to extract content-agnostic information and comprehensive speech features, respectively.
An orthogonal projection purification (OPP) module first isolates the content-agnostic component within the complex features and then eliminates it to obtain purified representations.
Additionally, to adequately extract content-agnostic information, we implement a supervision enhancement method that perturbs the speech input during training, accompanied by a consistency loss to constrain the representation, thereby enhancing the model's robustness and purification capability.

Notably, our method does not require transcriptions or additional annotations to accomplish the purification. As a result, it maintains higher flexibility and can be applied to unwritten languages that do not possess any transcription data.

We conduct experiments on the MuST-C and CoVoST-2 datasets, covering ten translation directions, in scenarios with and without transcriptions, as well as with additional MT data. The experimental results demonstrate the superiority of our method on all translation directions, and achieving preeminent performance without transcriptions.

\begin{figure}[t]
  \includegraphics[width=\columnwidth]{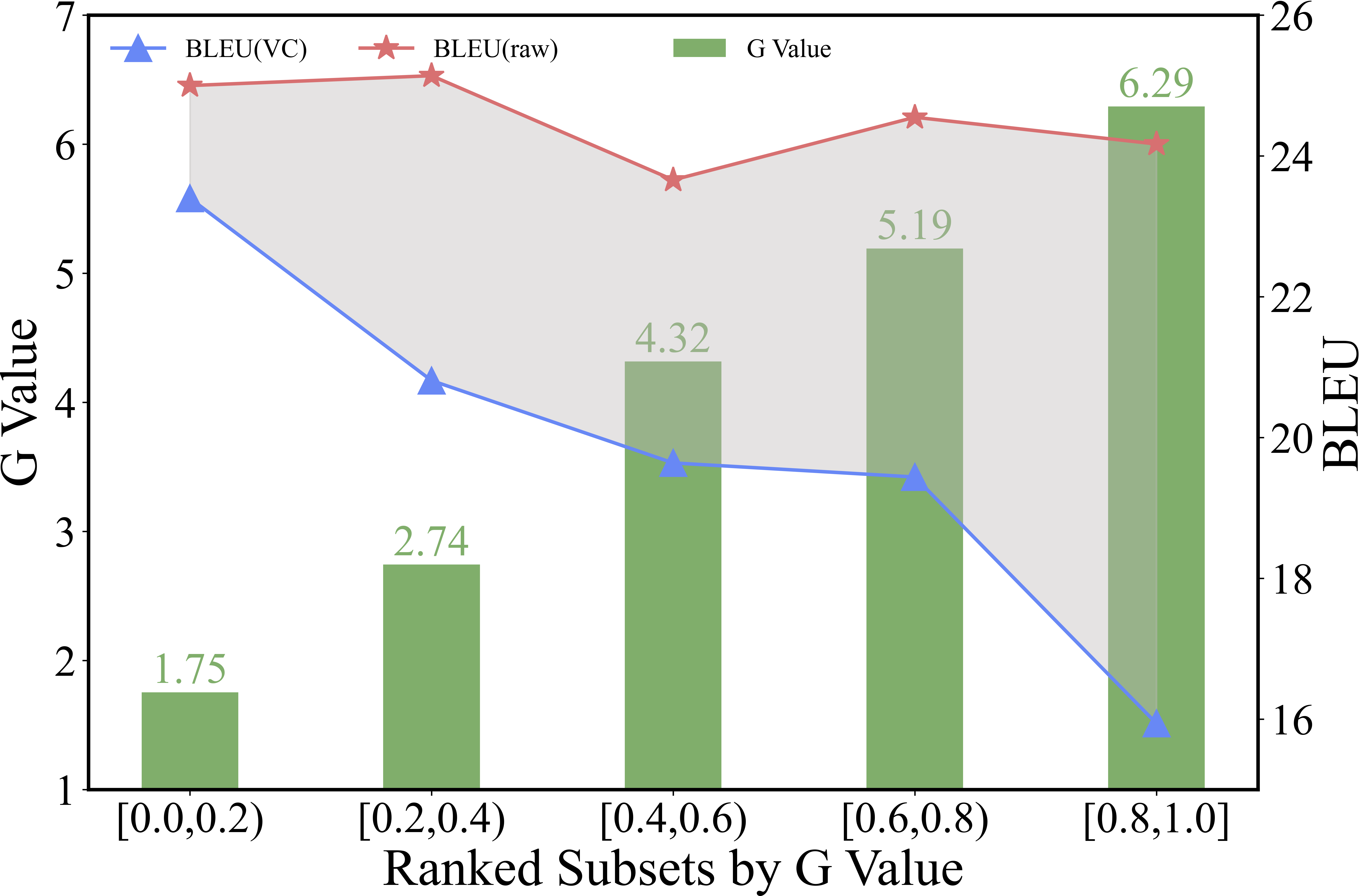}
  \caption{BLEU scores on MuST-C En-De dev subsets. \textbf{VC} and \textbf{raw} denote the BLEU scores are calculated with voice-converted audio $\tilde{\mathbf{s}}$ and raw audio $\mathbf{s}$, respectively. The Green bar denotes the G value.}
  \label{fig:pre_bleu}
\end{figure}
\begin{figure}[t]
  \includegraphics[width=\columnwidth]{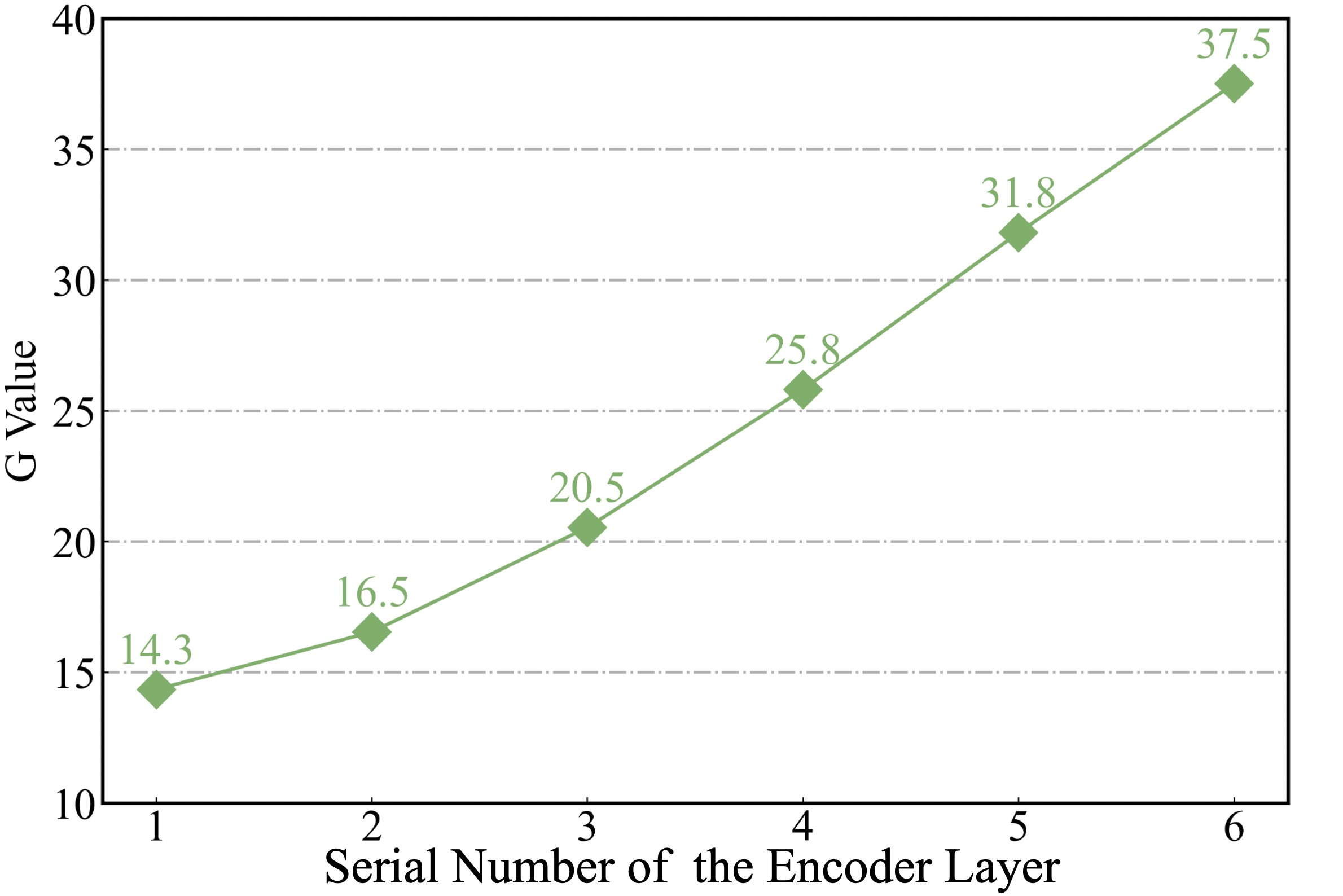}
  \caption{Averaged $G$ values of each T-Enc layer's outputs.}
  \label{fig:pre_layerwise_G}
\end{figure}
\begin{figure}[t]
  \includegraphics[width=\columnwidth]{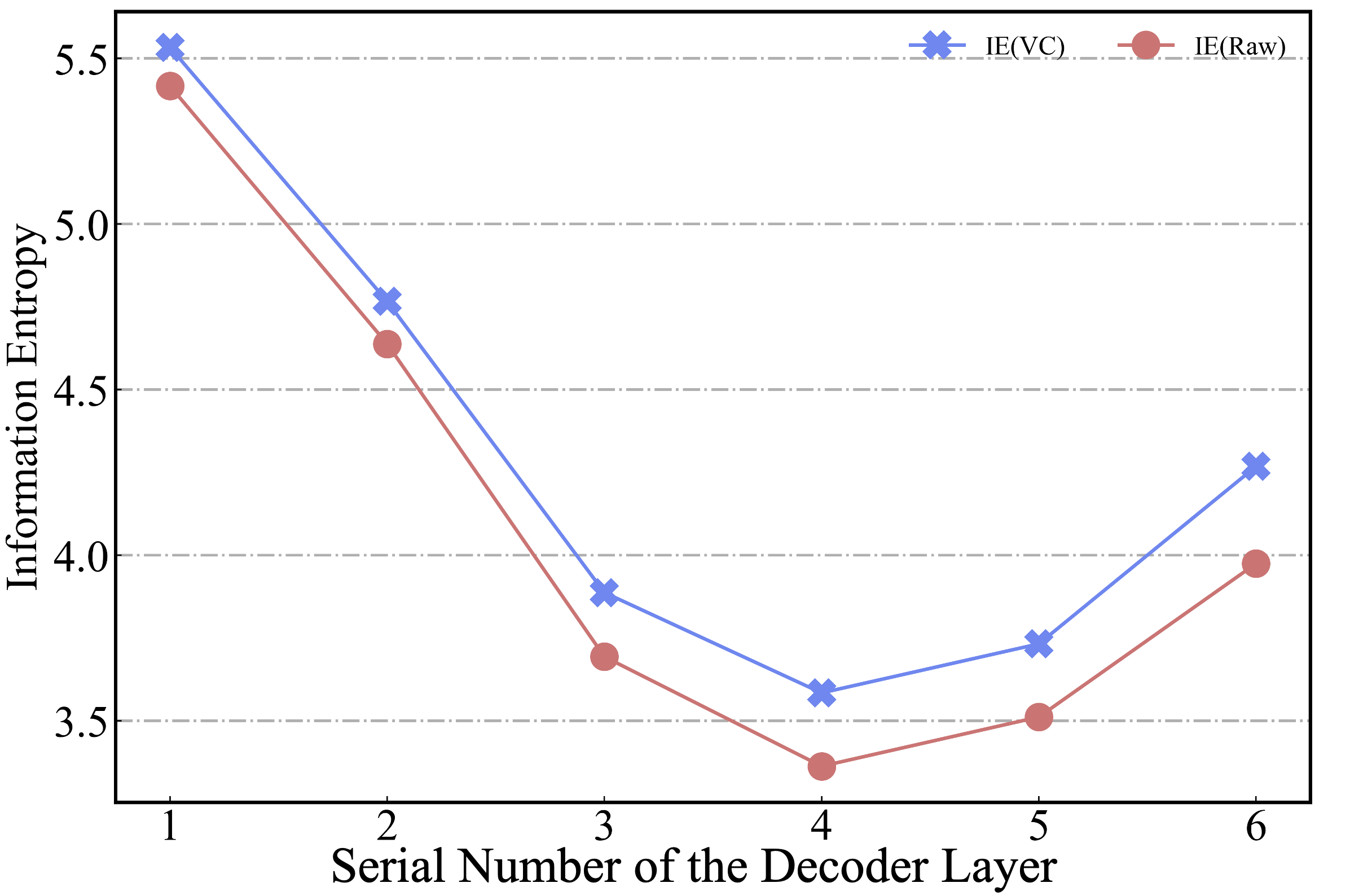}
  \caption{Averaged information entropy of cross-attention weights.}
  \label{fig:pre_cross_attn_IE}
\end{figure}
\section{Preliminary Analysis}
\label{sec:preliminary_analysis}
In this section, we examine the impact of content-agnostic perturbations on the ST model. Typically, an ST dataset that contains triplet data can be formed as $\mathcal{D}=\{(\mathbf{s}, \mathbf{x}, \mathbf{y})\}$, where $\mathbf{s}$, $\mathbf{x}$, $\mathbf{y}$ denote source speech, transcription, and translation, respectively.
To perturb in the content-agnostic aspects of speech while preserving the content-relevant information, we use a voice conversion (VC) system \cite{chou2019one} to modify the speaker’s information, transforming the source speech $\mathbf{s}$ into its perturbed version $\tilde{\mathbf{s}}$.
We conduct experiments based on XSTNet \citep{ye21_interspeech}, more experimental details are described in Appendix~\ref{appendix:preliminary}.
By feeding either $\mathbf{s}$ or $\tilde{\mathbf{s}}$ into the model, we measure the extent to which the model is influenced by content-agnostic perturbations, quantified by $G$, which is defined as the sentence-level L2 distance between the output representations of the textual encoder:
\begin{equation}\label{equation_G_value}
    G =\parallel {\rm \mathbf{Avg}}(f_{e}(\mathbf{s})) - {\rm \mathbf{Avg}}(f_{e}(\tilde{\mathbf{s}}))\parallel_2,
\end{equation}
where ${\rm \mathbf{Avg}(\cdot)}$ denotes average pooling on the temporal dimension, and $f_{e}(\cdot)$ means the corresponding output of textual encoder. A higher $G$ value indicates a greater impact on the model.

\noindent\textbf{Impact on Translation Quality} 
To demonstrate the correlation between the degree of perturbations and translation performance, we calculate $G$ for all samples in MuST-C~\citep{di-gangi-etal-2019-must} En-De dev set and divide the samples into five equal-sized subsets based on their $G$ values.
As shown in Figure~\ref{fig:pre_bleu}, when $\tilde{\mathbf{s}}$ is used as input we observe a significant decline in BLEU scores as the $G$ value increases; meanwhile, the BLEU gap (grey area in Figure~\ref{fig:pre_bleu}) widens.
These findings suggest that the model focuses excessively on context-agnostic information and is highly susceptible to perturbations.

\noindent\textbf{Impact on Textual Encoder} Furthermore, we investigate the response of textual encoder to perturbations by tracking fluctuations of $G$ value across each layer.
As illustrated in Figure~\ref{fig:pre_layerwise_G}, as the layers deepen, the $G$ value\footnote{Note that layer normalization is applied at the top of the final layer of the textual encoder, which accounts for the larger scale of $G$ values in Figure~\ref{fig:pre_layerwise_G} compared to Figure~\ref{fig:pre_bleu}.} consistently rises, indicating that the textual encoder fails to neutralize perturbations and cannot effectively extract content-relevant information.

\noindent\textbf{Impact on Decoder} 
To explore the relationship between representation perturbation and decoder performance, we compute the information entropy (IE) of cross-attention weights in each decoder layer, as shown in Figure~\ref{fig:pre_cross_attn_IE}.
Higher IE values are observed in every decoder layer when $\tilde{\mathbf{s}}$ input to the model, indicating greater uncertainty during the decoding process.
This phenomenon is more pronounced in deeper layers, contributing to performance degradation.

Based on the observations and analyses above, we can conclude that the model lacks robustness against content-agnostic information in speech, ultimately leading to performance degradation.
Inspired by these findings, we aim to extract and eliminate content-agnostic information from speech features to improve translation performance.

\section{Method}
\begin{figure}[htb]
    \centering
    \includegraphics[width=\columnwidth]{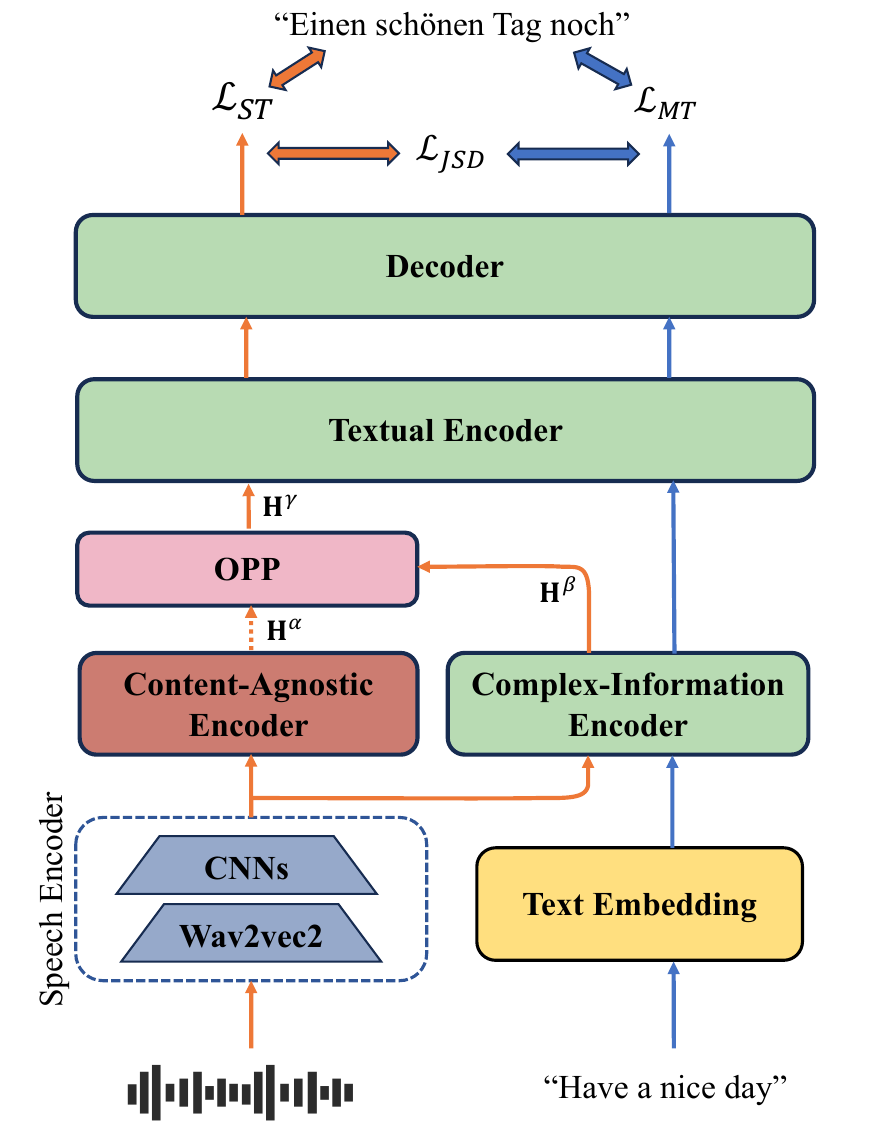}
    \caption{Overview of our proposed framework. The text embedding and MT forward path are deprecated during inference or training in the \textit{transcript-free} setting.}
    \label{fig:main_figure}
\end{figure}
\begin{figure}[ht]
    \centering
    \includegraphics[width=\columnwidth]{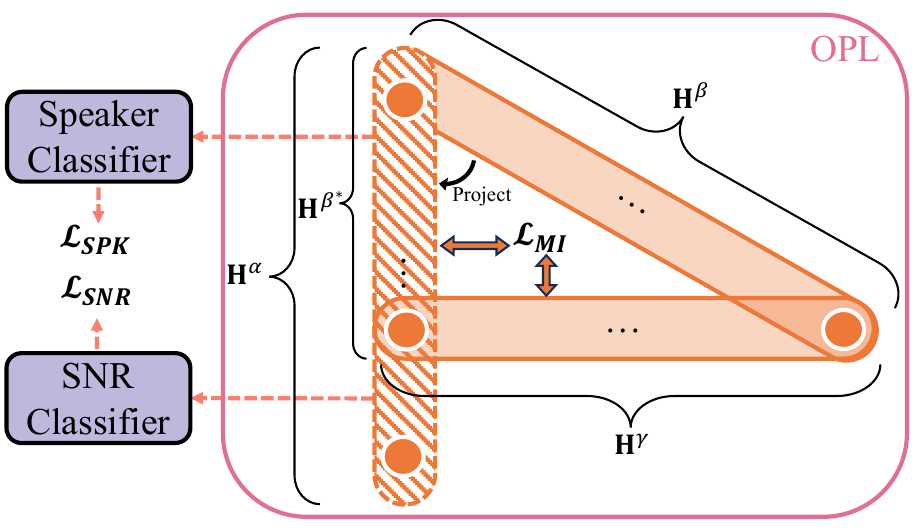}
    \caption{Diagram of OPP Module. It consists of two classifiers and an orthogonal projection layer.}
    \label{fig:OPD}
\end{figure}
In this section, we present our model architecture in Section~\ref{sec:model_architecture}, followed by a detailed explanation of the proposed \textbf{S}peech \textbf{R}epresentation \textbf{P}urification with \textbf{S}upervision \textbf{E}nhancement (SRPSE) method in Section~\ref{sec:ssrp}. An overview of our framework is illustrated in Figure~\ref{fig:main_figure}.

\subsection{Model Architecture}
\label{sec:model_architecture}
Our model primarily comprises six modules: the \textit{speech encoder} (S-Enc), the \textit{content-agnostic encoder} (CA-Enc), the \textit{complex-information encoder} (CI-Enc), the \textit{orthogonal projection purification} (OPP) module, the \textit{textual encoder} (T-Enc), and the \textit{decoder}.

\noindent\textbf{Speech Encoder} 
We adopt Wav2vec2.0 \texttt{base}~\citep{baevski2020wav2vec} to extract low-level features, and a two-layer 1D CNN with stride 2 to reduce the sequence length by a factor of 4.

\noindent\textbf{CA-Enc } \textbf{\&} \textbf{CI-Enc}
The CA-Enc and CI-Enc consist of $N^{\alpha}$ and $N^{\beta}$ Transformer encoder layers, respectively, which use the same configurations as the vanilla Transformer, except that pre-norm \citep{pmlr-v119-xiong20b} is applied for stable training.
The hyper-parameters $N^{\alpha}$ and $N^{\beta}$ are both set to 1.
The CA-Enc is expected to extract content-agnostic information, while the CI-Enc captures full information of speech.

\noindent\textbf{Orthogonal Projection Purification (OPP)}
As depicted in Figure~\ref{fig:OPD}, the OPP module mainly comprises three components: speaker classifier, signal-to-noise ratio (SNR) classifier, and orthogonal projection layer (OPL)~\citep{qin-etal-2020-feature}.
The speaker classifier and the SNR classifier predict speaker IDs and background noise levels , respectively, using the output representations of CA-Enc.
These classifiers are designed to provide supervisory information for CA-Enc.
The OPL is introduced to eliminate the content-agnostic aspects in complex features, thereby producing purified representations that are only relevant to the speech content.

\noindent\textbf{Textual Encoder}
With the same configurations as the CA-Enc and CI-Enc, the T-Enc further extracts the high-level semantic hidden representations of speech and text.

\noindent\textbf{Decoder}
We employ the base configuration as the vanilla Transformer decoder. It generates the translation sequences for ST or MT tasks. The corresponding translation objective is defined as:
\begin{equation}
    \label{eq:st}
    \mathcal{L}_{\rm ST} = - \sum_{(\mathbf{s},\mathbf{y}) \in \mathcal{D}} \log P(\mathbf{y} \mid \mathbf{s}),
\end{equation}
\begin{equation}
    \label{eq:mt}
    \mathcal{L}_{\rm MT} = - \sum_{(\mathbf{x},\mathbf{y})\in\mathcal{D}} \log P(\mathbf{y} \mid \mathbf{x}).
\end{equation}
Besides, we minimize the Jensen-Shannon Divergence (JSD) between MT and ST probability distributions to transfer knowledge from MT to ST:
\begin{equation}
    \begin{split}
    \mathcal{L}_{\rm JSD} = \sum_{(\mathbf{s}, \mathbf{x}, \mathbf{y}) \in \mathcal{D}} {\rm JSD} [P(\mathbf{y} \mid \mathbf{s}) \parallel P(\mathbf{y} \mid \mathbf{x})].
    \end{split}
\end{equation}

\subsection{Speech Representation Purification with Supervision Enhancement(SRPSE)}
\label{sec:ssrp}
As mentioned in Section~\ref{sec:preliminary_analysis}, we aim to purify the complex speech representations by dislodging the content-agnostic part.
Two major problems hamper us from achieving our goal: (1) Given the content-agnostic representation $\mathbf{H}^{\alpha}$ output by CA-Enc and the complex speech representation $\mathbf{H}^{\beta}$ output by CI-Enc, how do we produce the ideal purified speech representation $\mathbf{H}^{\gamma}$? (2) How do we ensure the $\mathbf{H}^{\alpha}$ truly includes adequate content-agnostic information?

\noindent\textbf{Orthogonal Projection Purification} 
To answer the first question we introduce the orthogonal projection layer (OPL) to eliminate the content-agnostic parts present in the complex features, producing a purified representation $\mathbf{H}^{\gamma}$ which is only relevant to the content.

Specifically, we first project the complex representation $\mathbf{H}^{\beta}$ extracted by the CI-Enc to the content-agnostic representation $\mathbf{H}^{\alpha}$ extracted by the CA-Enc to obtain $\mathbf{H}^{\beta^*}$:
\begin{equation}
    \label{eq:proj}
    \mathbf{H}^{\beta^*} = \frac{\mathbf{H}^{\beta} \cdot \mathbf{H}^{\alpha}}{\mid \mathbf{H}^{\alpha} \mid} \frac{\mathbf{H}^{\alpha}}{\mid \mathbf{H}^{\alpha} \mid}.
\end{equation}
This operation entails the mining of content-agnostic components within the complex features.
Then we project $\mathbf{H}^{\beta}$ to the orthogonal hyperplane of $\mathbf{H}^{\beta^*}$ to obtain $\mathbf{H}^{\gamma}$. In practice, this projection formed as:
\begin{equation}
    \mathbf{H}^{\gamma} = \mathbf{H}^{\beta} - \mathbf{H}^{\beta^*}.
\end{equation}
This process eradicates redundancy within complex features, yielding the purified speech representation. However, we expect there is no information overlapping between content-agnostic and purified representations, but the orthogonality of representations does not imply a complete absence of mutual information between them.
Thus we introduce vCLUB~\citep{pmlr-v119-cheng20b} to minimize mutual information upper bound between $\mathbf{H}^{\gamma}$ and $\mathbf{H}^{\beta^*}$:
% \begin{equation}
% \begin{split}
%    \mathcal{L}_{\rm MI} = \mathbb{E}_{p(\mathbf{H}^{\gamma}, \mathbf{H}^{\beta^*})}[\log q_{\theta}(\mathbf{H}^{\gamma} \mid \mathbf{H}^{\beta^*})] \\ -\mathbb{E}_{p(\mathbf{H}^{\gamma})}\mathbb{E}_{p(\mathbf{H}^{\beta^*})}[\log q_{\theta}(\mathbf{H}^{\gamma} \mid \mathbf{H}^{\beta^*})],
% \end{split}
% \end{equation}
\begin{equation}
\label{eq:mi}
\begin{split}
    \mathcal{L}_{\rm MI} & = \frac{1}{N} \sum_{i=1}^{N}[\frac{1}{T} \sum_{t=1}^{T} \log q_{\theta}(\mathbf{H}_i^{\gamma} \mid \mathbf{H}_i^{\beta^*}) \\ & - \frac{1}{N}\frac{1}{T} \sum_{j=1}^N \sum_{t=1}^T \log q_{\theta}(\mathbf{H}_j^{\gamma} \mid \mathbf{H}_i^{\beta^*})],
\end{split}
\end{equation}
where $q_{\theta}(\mathbf{H}^{\gamma} \mid \mathbf{H}^{\beta^*})$ serves as a variational approximation of posterior $p(\mathbf{H}^{\gamma} \mid \mathbf{H}^{\beta^*})$ with approximation network $\theta$. More details about the vCLUB and our implementation are elaborated in Apppendix~\ref{appendix:vclub}.
% Follows~\citet{NEURIPS2023_c89f0984}, we tailor it to suit our task, the mutual information loss is defined as:
% \begin{equation}
% \begin{split}
%     \mathcal{L}_{\rm MI} & = \frac{1}{N} \sum_{i=1}^{N}[\frac{1}{T} \sum_{t=1}^{T} \log q_{\theta}(\mathbf{H}_i^{\gamma} \mid \mathbf{H}_i^{\beta^*}) \\ & - \frac{1}{N}\frac{1}{T} \sum_{j=1}^N \sum_{t=1}^T \log q_{\theta}(\mathbf{H}_j^{\gamma} \mid \mathbf{H}_i^{\beta^*})],
% \end{split}
% \end{equation}
% where $N$ is the batch size and $T$ is sequence length.
% The approximation network $\theta$ and our model are optimized alternatively during training.
% All configurations of $\theta$ remain consistent with ~\citet{pmlr-v119-cheng20b}.

\noindent\textbf{Content-Agnostic Supervision Enhancement} For the second question, without stricter constraints on the CA-Enc, supervision signals generated by the mutual information minimization task may be insufficient.
Therefore, we employ speech perturbations to introduce richer supervision signals and further enhance the purification process.
We employ three perturbation policies: \textit{noise interference}, \textit{pitch shift} and \textit{time stretch}~\citep{park19e_interspeech}.
For each speech input, we randomly sample a signal-to-noise ratio $\varepsilon \in  \{5,10,20,50,+\infty \}$, a pitch shift step $\mu \in \{-1, 0,+1\}$ and a stretch rate $\tau \in \{0.8, 0.9, 1.0, 1.1, 1.2\}$.
Then a transformation defined by these three factors is applied to each sample using torchaudio toolkit \citep{yang2021torchaudio}:
\begin{equation}
    \tilde{\mathbf{s}} = f_{(\varepsilon, \mu, \tau)}(\mathbf{s}),
\end{equation}
when $\varepsilon = +\infty$, $\mu=0$, and $\tau=1.0$, it denotes the policy is not applied.

The $\tilde{\mathbf{s}}$ also forward in S-Enc, CA-Enc, CI-Enc, and OPP module.
With speaker IDs and $\varepsilon$ serving as content-agnostic supervision signals, we can now regularize the CA-Enc by these two classifiers mentioned in (Section~\ref{sec:model_architecture}) with:
\begin{equation}
\begin{split}
    \mathcal{L}_{\rm SPK} & = - \frac{1}{2}{\textstyle \sum_{i=1}^{\mid \mathcal{D} \mid}} [\log P(\mathbf{spk}_i \mid \mathbf{H}^{\alpha}) \\ & + \log P(\mathbf{spk}_i \mid \widetilde{\mathbf{H}}^{\alpha})],
\end{split}
\end{equation}
\begin{equation}
\begin{split}
    \mathcal{L}_{\rm SNR} & = - \frac{1}{2} [{\textstyle \sum_{i=1}^{\mid \mathcal{D} \mid}} \log P(\varepsilon_i \mid \mathbf{H}^{\alpha}) \\ & + {\textstyle \sum_{j=1}^{\mid \mathcal{D} \mid}} \log P(\varepsilon_j \mid \widetilde{\mathbf{H}}^{\alpha})],
\end{split}
\end{equation}
where $\widetilde{\mathbf{H}}^{\alpha}$ is the counterpart of $\mathbf{H}^{\alpha}$ when $\tilde{s}$ input to the model. We don't utilize $\mu$ and $\tau$ explicitly, but incorporating these two perturbations enhances the difficulty of predicting speaker IDs, providing sterner regularization to CA-Enc.

Theoretically, if SRPSE is capable of filtering out all content-agnostic information, it should generate a similar representation regardless of the $\mathbf{s}$ or $\tilde{\mathbf{s}}$ serves as input to our model.
Therefore we anticipate a higher degree of proximity between $\mathbf{H}^{\gamma}$ and its counterpart $\widetilde{\mathbf{H}}^{\gamma}$. We average $\mathbf{H}^{\gamma}$ and $\widetilde{\mathbf{H}}^{\gamma}$ on temporal dimension to get sentence-level representation and employ a consistency loss to bring them together:
\begin{equation}
\begin{split}
    \mathcal{L}_{\rm CONSIS} = \sum^{\mid \mathcal{D} \mid} \parallel {\rm \mathbf{Avg}}({\mathbf{H}^{\gamma}}) - {\rm \mathbf{Avg}}(\widetilde{\mathbf{H}}^{\gamma}) \parallel_2.
\end{split}
\end{equation}

The overall training objectives of transcript-free setting and multi-task setting are as follows:
\begin{equation}
\label{eq:trans_free}
\begin{split}
    \mathcal{L}_{\rm TF} & = \mathcal{L}_{\rm ST} + \mathcal{L}_{\rm SPK} + \mathcal{L}_{\rm SNR} \\ & + \lambda_1 \mathcal{L}_{\rm CONSIS} + \lambda_2 \mathcal{L}_{\rm MI},
\end{split}
\end{equation}
\begin{equation}
\label{eq:mtl}
    \mathcal{L}_{\rm MTL} = \mathcal{L}_{\rm TF} + \mathcal{L}_{\rm MT} + \mathcal{L}_{\rm JSD},
\end{equation}
where $\lambda_1$ and $\lambda_2$ are hyper-parameters.

\section{Experiments}
\subsection{Experimental Setup}
\noindent\textbf{Datasets} We conduct experiments on MuST-C~\citep{di-gangi-etal-2019-must} and CoVoST-2~\citep{wang2020covost} datasets. MuST-C is a one-to-many ST dataset, covering pairs from English to Dutch (Nl), French (Fr), German (De), Italian (It), Portuguese (Pt), Romanian (Ro), Russian (Ru), and Spanish (Es). CoVoST-2 is a large and diversified multilingual ST corpus, we experiment in the German-English and French-English directions. Both of these datasets comprise triplet data sources: speech, transcription, and translation, which are meticulously aligned at the sentence level. For a fair and comprehensive comparison, we follow~\cite{du2022regularizing, zhou2023cmot}, the WMT16 En-De, WMT14 En-Fr, and WMT13 En-Es serve as external data for German, French, and Spanish translation respectively. The detailed statistics for all datasets are shown in Appendix~\ref{appendix:datasets}.

\noindent\textbf{Training settings}
There are three settings for speech translation tasks: \textit{transcript-free}, \textit{multi-task}, and \textit{expanded}.
For \textit{transcript-free} setting, only the $(\mathbf{s}, \mathbf{y})$ pairs are used to train our model, and the training objective is Equation~\ref{eq:trans_free}. 
For \textit{multi-task} setting, we use $(\mathbf{s}, \mathbf{x}, \mathbf{y})$ triplets with Equation~\ref{eq:mtl}.
For \textit{expanded} setting, we first pre-train the corresponding components with the external MT dataset then fine-tune our model with progressive training~\cite{ye2021end} on MuST-C triplets.

\noindent\textbf{Experiment Details}
\label{sec:exp_details}
The implementation of our model is based on fairseq\footnote{\url{https://github.com/facebookresearch/fairseq}}~\citep{ott2019fairseq}. The hyper-parameters $\lambda_1$, $\lambda_2$, $N^{\alpha}$, and $N^{\beta}$ are set to $1.0$, $0.01$, $1$ , and $1$, respectively. The textual encoder and the decoder consist of 5 and 6 layers, respectively. We report case-sensitive detokenized BLEU scores using SacreBLEU~\citep{post-2018-call} in our main results, and additionally present ChrF++~\citep{popovic-2017-chrf} and COMET~\citep{rei-etal-2022-comet} scores in our ablation study and analysis. Appendix ~\ref{appendix:Experiment} shows more implementation details and explanations for the baselines. Detailed hyper-parameter selection experiments are also provided in Appendix~\ref{appendix:hyper-parameter}.
\begin{table*}[htb]
    \small
    \normalsize
    \centering
    \setlength{\tabcolsep}{1.7mm}{
    \scalebox{0.95}{
    \resizebox{\textwidth}{!}{
        \begin{tabular}{lccccccccc}
            \hline
            \toprule
            \textbf{Models} & \textbf{En-De} & \textbf{En-Fr} & \textbf{En-Ru} & \textbf{En-Es}  & \textbf{En-It}  & \textbf{En-Nl} & \textbf{En-Pt} & \textbf{En-Ro}& \textbf{Avg.} \\ \midrule
            \multicolumn{10}{c}{Training in \textit{transcript-free} setting} \\ \midrule
            Fairseq ST \cite{wang2020fairseq}  & 22.7 & 32.9 & 15.3 & 27.2 & 22.7 & 27.3 & 28.1 & 21.9 & 24.8 \\
            Revisit ST \cite{pmlr-v162-zhang22i} & 23.0 &   33.5 &15.6 & 28.0 & 23.5 & - & - &- & -\\
            W2V2-Transformer\cite{fang2022stemm}    & 24.1 & 35.0 & 16.3 & 29.4  & 24.8 &28.9 &30.0 &23.1& 26.5 \\ 
            CCSRD \cite{zhao-etal-2023-ccsrd} &25.4 &35.8 &16.8 & 30.2 &25.8& -& -& -& - \\
            % DUB (Base) \cite{zhang2023dub}$\dag$ &25.8 &34.7 &-& 30.2 &-& -& -& -& - \\
            DUB (Large) \cite{zhang2023dub}$\dag$ & 26.2 &35.3 &-& 30.4 &-& -& -& -& -\\
            BT4ST \cite{fang-feng-2023-back}$\dag$ &\textbf{26.6} &\textbf{36.9} &-& \textbf{31.2} &-& -& -& -& - \\
            \midrule
            \textbf{SRPSE}   & 26.2* & 36.5* & \textbf{17.6}* & \textbf{31.2}* & \textbf{26.1}* & \textbf{30.4}* & \textbf{31.9}* & \textbf{24.6}* & \textbf{28.0}* \\ \midrule
            \multicolumn{10}{c}{Training in \textit{multi-task} setting} \\ \midrule
            Memory-ST \cite{yuan2024memory}  &23.2 & 33.5 & - & 28.6  & 23.9 &27.6 &28.7 & - & - \\ 
            % SATE \cite{xu2021stacked} & 25.2 & - & - & - & - & - & - & - & -\\
            % TDA \cite{du2022regularizing} & 25.4 & 36.1 & 16.4 & 29.6  & 25.1 & 29.6 &31.1 &23.9 &27.2\\
            XSTNet \cite{ye2021end} & 25.5 & 36.0 & 16.9 & 29.6  & 25.5 &30.0 &31.3 &25.1 &27.5\\ 
            STEMM \cite{fang2022stemm} & 25.6 & 36.1 & 17.1 & 30.3  & 25.6 &30.1 &31.0 &24.3 &27.5\\
            ConST \cite{ye2022cross} & 25.7 & 36.8 & 17.3 & 30.4 & 26.3 & 30.6 & 32.0 &24.8 &28.0\\
            MCTN \cite{zhou2024multitask} &25.9 &36.1 &17.1 &30.3 &25.7& -& -& -& -\\

            Siamese-PT \cite{pmlr-v202-le23a} &26.2 &36.9 &16.8 & 29.8 &25.9& 29.8 & 32.1& 24.8& 27.8\\
            CCSRD \cite{zhao-etal-2023-ccsrd} &26.1 &37.1 &17.8 & 31.0 &26.4& -& -& -& -\\
            M$^{3}$ST \cite{cheng2023m} &26.4 &37.2 &\textbf{18.3} &31.0 & 26.6& 30.9& \textbf{32.8}& 25.4& 28.6\\
            CMOT \cite{zhou2023cmot} &\textbf{27.0} &37.3 &17.9 &31.1 & 26.9& 31.2 &32.7 &25.3& 28.7 \\
            
            \midrule
            \textbf{SRPSE} & 26.9* & \textbf{37.4}* & \textbf{18.3}* & \textbf{31.4}* & \textbf{27.0}* & \textbf{31.4}* & \textbf{32.8}* & \textbf{25.5}* & \textbf{28.8}* \\
            \bottomrule
            \hline
        \end{tabular}}}}
    \caption{BLEU scores on MuST-C tst-COMMON set under \textit{transcript-free} setting and \textit{multi-task} setting. $\dag$ indicates external target-side MT data was used during training. * denotes the improvements over the W2V2-Transformer baseline in transcript-free setting and XSTNet baseline in multitask-setting is statistically significant ($p < 0.01$).}
    \label{tab:main_results_bleu}
\end{table*}
\begin{table}[htb]
    \small
    \normalsize
    \centering
    \setlength{\tabcolsep}{1mm}{
    \scalebox{1.0}{
        \resizebox{\columnwidth}{!}{
        \begin{tabular}{lcc}
            \hline
            \toprule
            \textbf{Models} & \textbf{Fr-En} & \textbf{De-En} \\ \midrule
            Transformer-ST \cite{wang2020covost} & 26.3 & 17.1 \\
            Revisit ST \cite{pmlr-v162-zhang22i} & 26.9 & 14.1 \\
            U2TT (Large) \cite{zhang2023dub} & 27.4 & 16.7 \\ 
            DUB (Large) \cite{zhang2023dub}$\dag$ & \textbf{29.5} & 19.5 \\
            \midrule
            \textbf{SRPSE} & 29.3 & \textbf{21.4} \\
            \bottomrule
            \hline
        \end{tabular}
        }
    }}
    \caption{BLEU scores on CoVoST-2 De-En and Fr-En test sets under \textit{transcript-free} setting. $\dag$ indicates external targe-side MT data was used during training.}
    \label{tab:covost_results_bleu}
\end{table}
\begin{table}[t]
    \centering
    \setlength{\tabcolsep}{1mm}{
        \resizebox{\columnwidth}{!}{
        \begin{tabular}{lccc}
            \hline
            \toprule
            \textbf{Models} & \textbf{En-De} & \textbf{En-Fr} & \textbf{En-Es} \\ \midrule
            % AdaTranS\cite{zeng-etal-2023-adatrans}           & Y    & 26.7  & 37.4  & -  & - \\ 
            W2V2-Transformer \cite{fang2022stemm}   & 26.9 & 36.6 & 30.0\\ 
            TDA \cite{du2022regularizing}    &  27.1   & -  & -  \\
            Chimera \cite{han2021learning} & 27.1 & 35.6 & 30.6\\
            SATE \cite{xu2021stacked} & 28.1 & - & -\\
            STEMM \cite{fang2022stemm}  & 28.7 & 37.4 & 31.0 \\
            XSTNet \cite{ye2021end}  & 27.8 & 38.0 & 30.8 \\
            ConST \cite{ye2022cross}& 28.3 &  38.3 & 32.0 \\ 
            CMOT\cite{zhou2023cmot} & 29.0 &  39.5 &32.8 \\ 
            \midrule
            \textbf{SRPSE} & \textbf{29.2}* & \textbf{39.9}* & \textbf{33.0}* \\
            \bottomrule
            \hline
        \end{tabular}
        }
    }
    \caption{BLEU scores on MuST-C tst-COMMON set with external training data (\textit{expended} setting). * means the improvements over XSTNet are statistically significant ($p < 0.01$).}
    \label{tab:extended_results_bleu}
\end{table}
\subsection{Main Results}
\textbf{Comparison with End-to-End Baselines} 
The main results on the MuST-C and CoVoST-2 datasets are presented in Table~\ref{tab:main_results_bleu} and Table~\ref{tab:covost_results_bleu}, respectively.
In the \textit{transcript-free} setting, our model achieves distinguished performance, and significantly outperforms W2V2-Transformer~\citep{fang2022stemm} by an average of 1.5 BLEU scores. It attains either superior or comparable performance on MuST-C and CoVoST-2 with fewer parameters and less training data\footnote{DUB (larger) has a larger model size than ours (260M vs. 160M), BT4ST employs multiple models for back translation, and both of them utilize additional target-side MT data, whereas our model only uses speech-translation pairs.}, compared to our strongest baselines, DUB (Large)~\citep{zhang2023dub} and BT4ST~\citep{fang-feng-2023-back}.  
In the \textit{multi-task} setting, our model exceeds XSTNet~\citep{ye2021end} on MuST-C by an average of 1.3 BLEU scores and surpasses our strongest baseline, CMOT~\citep{zhou2023cmot}.
As shown in Table~\ref{tab:extended_results_bleu}, with the introduction of external MT data, our model also gains an average of 1.8 BLEU scores improvement compared with XSTnet and outperforms CMOT slightly.
These gains verify the effectiveness of our approach.

Among all baselines, CCSRD~\citep{zhao-etal-2023-ccsrd} aim to address similar issues, they chose to encode the speech representation into two components directly and the cyclic reconstruction is a sophisticated decoupling approach.
Unlike CCSRD, our approach focuses on extracting and filtering out the redundant parts in speech representations.

\noindent\textbf{Comparison with Cascaded Baselines}
Table~\ref{tab:cascaded_results} illustrates the performance of our model compared to several cascaded baselines. 
Among these, the \textbf{Cascade} refers to our implementation of a cascade system. The ASR part is trained on a mixture of LibriSpeech \cite{librispeech} and MuST-C data, and the MT part is trained on external MT and MuST-C data. 
The statistics denote SRPSE significantly outperforms all cascade baselines.

\begin{table}[t]
    \small
    \normalsize
    \centering
    % \resizebox{\linewidth}{!}{
        \begin{tabular}{lcc}
            \hline
            \toprule
            \textbf{Models}  & \textbf{En-De} & \textbf{En-Es} \\ \midrule
                   Espnet~\citep{inaguma2020espnet}   &  23.6 & 28.7     \\
            \citet{ye2021end}    & 25.2  & -    \\ 
            \citet{xu2021stacked}  & 28.1 & -  \\
            Cascade  & 26.8 &30.3  \\ 
            \midrule
            \textbf{SRPSE}     & \textbf{29.2}* & \textbf{33.0}* \\
            \bottomrule
            \hline
        \end{tabular}
    \caption{Our method versus the cascaded models on MuST-C En-De and En-Es tst-COMMON set. \textbf{Cascade} is a strong cascaded system we implemented. * mean the improvements over the cascaded baseline are statistically significant ($p < 0.01$).}
    \label{tab:cascaded_results}
\end{table}

\subsection{Ablation Study}
To evaluate the contribution of each training objective, we progressively eliminate them, and the results are shown in Table~\ref{tab:ablation_objectives}. 
First, we remove $\mathcal{L}_{\rm SPK}$ solely, resulting in a slight drop in both BLEU and ChrF++ by 0.2 points, and COMET drop by 0.5 points, indicating that our method does not heavily rely on speaker annotations from the ST dataset.
In Exp.\uppercase\expandafter{\romannumeral4}, when $\mathcal{L}_{\rm CONSIS}$, $\mathcal{L}_{\rm SPK}$, and $\mathcal{L}_{\rm SNR}$ are removed, BLEU scores decrease by 0.5 points, indicating the positive impact of our supervision enhancement strategy.
Further removing the $\mathcal{L}_{\rm MI}$ and deleting CA-Enc and OPP Module in Exp.\uppercase\expandafter{\romannumeral5}, we observed a decrease of 0.3 BLEU scores, proving that the constraint of mutual information is effective. 
In Exp.\uppercase\expandafter{\romannumeral6}, with the absence of $\mathcal{L}_{\rm JSD}$, the BLEU scores dropped by $0.4$, highlighting the significant impact of knowledge transfer.
\begin{table}[tb]
    \small
    \normalsize
    \centering
    \setlength{\tabcolsep}{1mm}{
        \resizebox{\columnwidth}{!}{
        \begin{tabular}{c|ccccc|ccc}
            \hline
            \toprule
              \#Exp.& $\mathcal{L}_{\rm CONSIS}$ & $\mathcal{L}_{\rm SNR}$ & $\mathcal{L}_{\rm SPK}$ & $\mathcal{L}_{\rm MI}$ & $\mathcal{L}_{\rm JSD}$ & BLEU & ChrF++ & COMET\\
             \midrule
             \uppercase\expandafter{\romannumeral1} & \checkmark & \checkmark & \checkmark & \checkmark & \checkmark & 26.9 & 54.1 & 75.2\\
             \uppercase\expandafter{\romannumeral2} & \checkmark & \checkmark & $\times$  & \checkmark & \checkmark & 26.7 & 53.9 & 74.7\\
             \uppercase\expandafter{\romannumeral3} & $\times$ & \checkmark & \checkmark & \checkmark & \checkmark & 26.7 & 54.0 & 74.9\\
             \uppercase\expandafter{\romannumeral4} & $\times$ & $\times$ & $\times$ & \checkmark & \checkmark & 26.4 & 53.6 & 74.1\\
             \uppercase\expandafter{\romannumeral5} & $\times$ & $\times$ & $\times$ & $\times$ & \checkmark & 26.1 & 53.3 & 73.7\\
             \uppercase\expandafter{\romannumeral6} & $\times$ & $\times$ & $\times$ & $\times$ & $\times$ & 25.7 &52.8 & 73.0 \\
            \bottomrule
            \hline
        \end{tabular}}
    }
    \caption{Ablation on training objectives under \textit{multi-task} setting. BLEU, ChrF++, and COMET scores are reported on MuST-C En-De tst-COMMON set.}
    \label{tab:ablation_objectives}
\end{table}
\begin{figure}[t]
    \centering
    \includegraphics[width=\columnwidth]{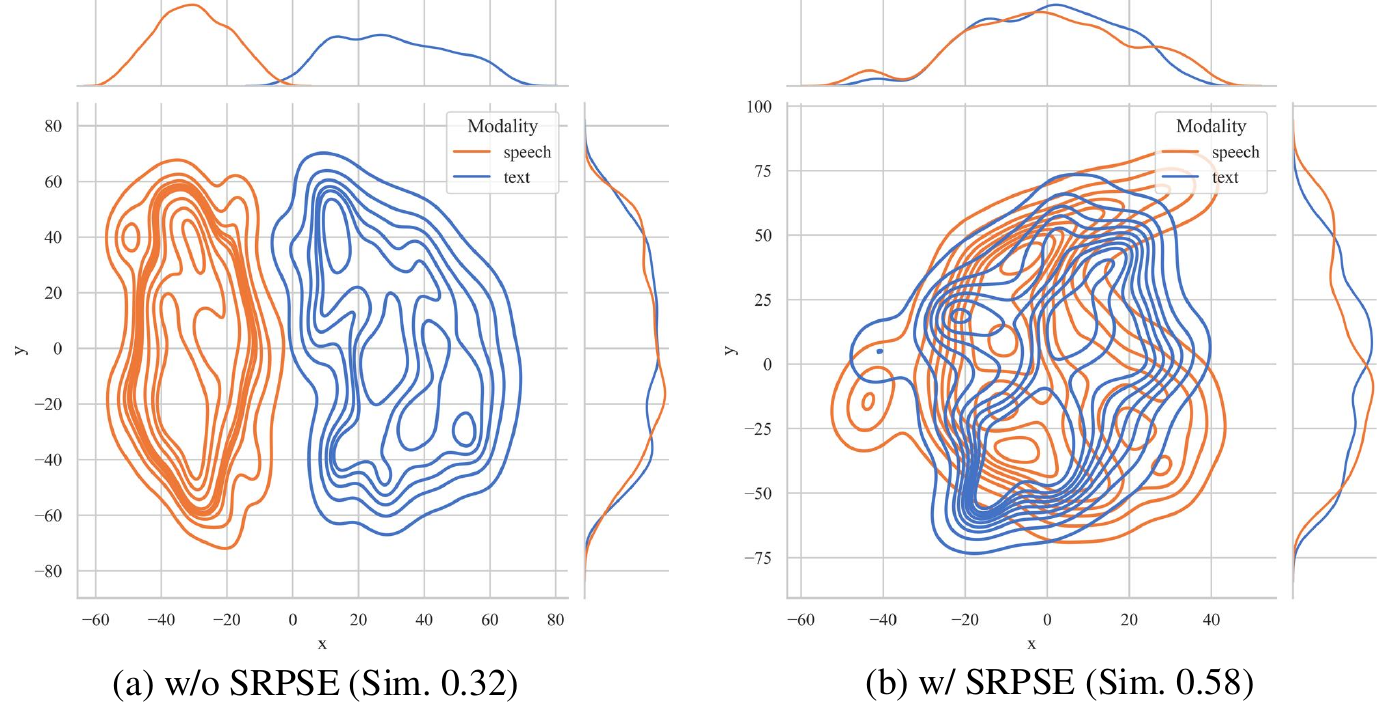}
    \caption{Bivariate KDE contour plot of speech and text representations on MuST-C En-De tst-COMMON set. Yellow and blue lines are speech and text representations respectively. T-SNE is utilized to reduce dimension to 2D. Sim. denotes the cosine similarity between these two representations. (a) The same configurations as that in Table~\ref{tab:ablation_objectives} Exp.\uppercase\expandafter{\romannumeral4}. (b) Our SRPSE.}
    \label{fig:tsne}
\end{figure}
\section{Analysis}

\subsection{Can SRPSE Purify Speech Representation?}
To assess the effectiveness of our approach in purifying speech representations,
we extract text and speech representations from T-Enc input and visualize them using t-SNE~\citep{van2008visualizing}.
Additionally, we calculate the average cosine similarity between cross-modal representations.
Figure~\ref{fig:tsne} is the bivariate kernel density estimation (KDE) plot, where greater overlap indicates more similar representation distributions. 
Without SRPSE, speech and text representations are clearly separated, with a relatively low average cosine similarity of $0.32$.
When SRPSE is applied, the representations are brought significantly closer, leading to a higher cosine
similarity of $0.58$.
Such phenomena suggest our approach can generate purified speech representation that contains more content-relevant information, and accounts for higher consistency with its text counterpart.

\subsection{Is SRPSE Robust to Content-agnostic Perturbations?}
\begin{figure}[t]
  \includegraphics[width=\columnwidth]{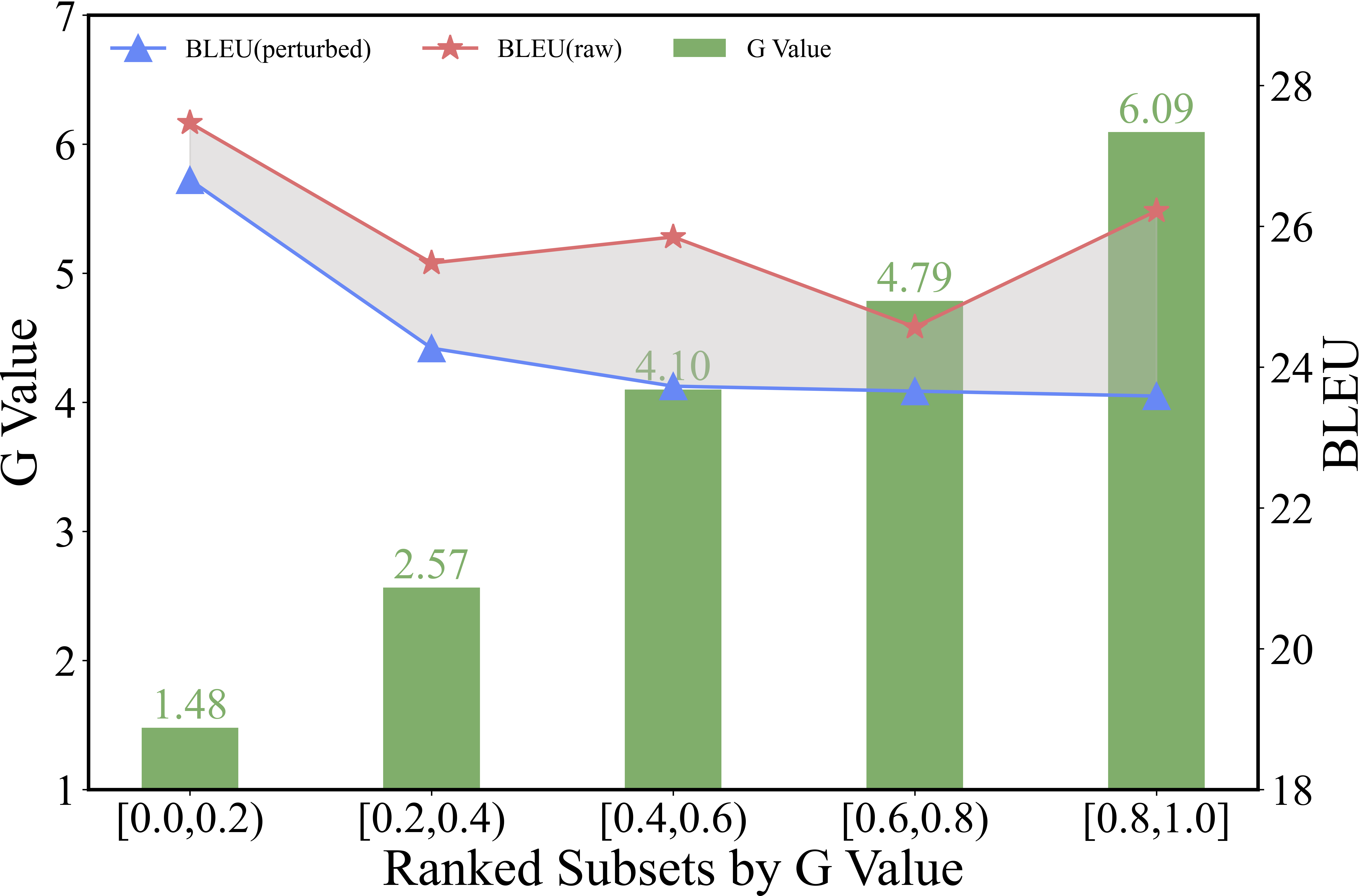}
  \caption{BLEU scores on MuST-C En-De dev subsets with our SRPSE. \textbf{Perturbed} and \textbf{raw} denote the BLEU scores are calculated with perturbed audio $\tilde{\mathbf{s}}$ and raw audio $\mathbf{s}$ respectively. The Green bar denotes the $G$ value.}
  \label{fig:post_bleu}
\end{figure}
We conduct the same experiment as in Section~\ref{sec:preliminary_analysis}, using voice conversion to perturb the speech input to assess the robustness of our model.
Figure~\ref{fig:post_bleu} illustrates the trend in BLEU scores as the $G$ value increases.
The averaged $G$ value across 5 subsets in Figure~\ref{fig:pre_bleu} is $4.05$, while our model is $3.8$, suggesting the representations of our model have higher stability.
Notably, the BLEU gap (grey area) is greatly reduced compared to Figure~\ref{fig:pre_bleu}.
As evident from these experimental findings, SRPSE achieves better performance under perturbations, confirming that our SRPSE enhances robustness.

\subsection{What's the Difference Between our Supervision Enhancement and Data Augmentation?}
To validate the effectiveness of our architecture and differentiate our approach from the conventional data augmentation method, we conduct experiments for further analysis.
We re-implement the Exp.\uppercase\expandafter{\romannumeral5} in Table~\ref{tab:ablation_objectives}, but using the perturbation method described in Section~\ref{sec:ssrp} as a data augmentation method, resulting in BLEU scores of $26.12$.
Despite augmenting the audio, there was only a negligible improvement compared to Exp.\uppercase\expandafter{\romannumeral5}.
This clarifies that the performance gains of our method stem from a delicately designed model structure rather than merely expanding the training data.

\subsection{What's the Additional Computational Cost Associated with the Introduction of New Modules?}
To evaluate the efficiency of our pipeline, we conduct experiments on MuST-C~\citep{di-gangi-etal-2019-must} En-De tst-COMMOM set. We set both beam size and batch size to 1 and performed inference on this set for 10 runs. For the baseline W2V2-Transformer~\citep{fang2022stemm}, the average time cost is $421.83$ seconds and the average tps (tokens per second) is $166.51$. In comparison, our model has an average time cost of $442.78$ seconds and a tps of $156.72$. This represents an approximately $5\%$ increase in inference time and a $6\%$ decrease in tps compared to the baseline. These results demonstrate that the additional modules introduce minimal computational overhead.

\section{Related Work}
% \subsection{End-to-End Speech Translation}
Training an end-to-end ST model that does not produce intermediate transcriptions is no easy job because of the modality gap and the scarcity of \textit{speech-transcription-translation} supervised data. To address these issues, many techniques have been used, including pretraining \cite{pino2020self, alinejad2020effectively, dong2021listen, xu2021stacked}, multi-task learning \cite{tang2021improving, ye2021end, vydana2021jointly}, data augmentation \cite{lam2022sample, mi2022improving}, meta-learning \cite{indurthi2020end}, and cross-modal alignment~\cite{han2021learning, xu2021stacked, ye2022cross, fang2022stemm}.

While most research chose to migrate the translation ability from MT to ST by designing exquisite model architectures or training procedures, few studies have investigated the correlation between speech characteristics and translation performance.
\citet{10447494} noticed the intrinsic modal differences and proposed to align the representation space rather than individual sample pairs, avoiding directly modifying the speech representation.
\citet{zhao-etal-2023-ccsrd} tackled this issue more straightforwardly, they proposed to decompose speech representation into content and non-content representation via disentanglement representation learning. 
% Distinct from prior efforts, our study emphasizes the extraction of the content-agnostic part of speech representations, coupled with a purification framework to eliminate it, ultimately elevating translation quality.

Representation purification aims to decompose various components behind data and utilize partial components to improve specific tasks, which is used extensively across numerous fields~\citep{qin-etal-2020-feature, s23167282, li2023unified, 10008072, 9682736}.
In text-to-speech (TTS) tasks, there has been a trend to decouple multiple acoustic features from speech to generate expressive speech. \citet{pmlr-v80-skerry-ryan18a}, and \citet{lee2021styler} proposed to disentangle prosody information for synthesizing high-quality audio.
\citet{pmlr-v119-qian20a}, \citet{yang22f_interspeech} and \citet{9747763} suggested that disentangling more aspects of speech could boost the performance of TTS.
In this paper, we conduct comprehensive experiments to investigate the correlation between speech translation quality and various speech components.
Based on our experiment results, our method distinct from prior efforts, emphasizes the extraction of the content-agnostic part of speech representations, coupled with a purification framework to eliminate it, ultimately elevating translation quality.

\section{Conclusion}
In this paper, we propose SRPSE, an ST framework that purifies speech representation by eliminating content-agnostic information.
The experimental results demonstrate the validity of the proposed framework under three training settings.
In-depth analyses demonstrate that SRPSE successfully purifies the speech representation and achieves higher robustness against content-agnostic perturbations.

\section*{Limitations}
Although the proposed method facilitates ST to purify speech representation and obtains significant improvements over previous methods, it still has some limitations:
% (1) There are too many content-agnostic factors in speech, and our supervision enhancement approach may not be diverse enough to cover all of them.
(1) There are too many content-agnostic factors in speech, only some of which are explored in this paper.
(2) The content-agnostic factors extraction granularity is not fine enough, some of these factors could be also used to improve ST.
(3) Whether our method can still be combined with multi-modal large language models to further improve the translation performance is unclear.
We leave these to our future exploration.

\section*{Acknowledgments}
This work is supported by the National Science and Technology Major Project (Grant No. 2022ZD0116101), the Key Support Project of NSFC-Liaoning Joint Foundation (Grant No. U1908216), and the public technology service platform project of Xiamen City (No. 3502Z20231043).

\bibliography{custom}

\begin{thebibliography}{66}
\providecommand{\natexlab}[1]{#1}

\bibitem[{Alinejad and Sarkar(2020)}]{alinejad2020effectively}
Ashkan Alinejad and Anoop Sarkar. 2020.
\newblock Effectively pretraining a speech translation decoder with machine translation data.
\newblock In \emph{Proc. of EMNLP}, pages 8014--8020.

\bibitem[{Ardila et~al.(2020)Ardila, Branson, Davis, Henretty, Kohler, Meyer, Morais, Saunders, Tyers, and Weber}]{commonvoice:2020}
R.~Ardila, M.~Branson, K.~Davis, M.~Henretty, M.~Kohler, J.~Meyer, R.~Morais, L.~Saunders, F.~M. Tyers, and G.~Weber. 2020.
\newblock Common voice: A massively-multilingual speech corpus.
\newblock In \emph{Proceedings of the 12th Conference on Language Resources and Evaluation (LREC 2020)}, pages 4211--4215.

\bibitem[{Baevski et~al.(2020)Baevski, Zhou, Mohamed, and Auli}]{baevski2020wav2vec}
Alexei Baevski, Yuhao Zhou, Abdelrahman Mohamed, and Michael Auli. 2020.
\newblock wav2vec 2.0: A framework for self-supervised learning of speech representations.
\newblock \emph{Advances in Neural Information Processing Systems}, 33:12449--12460.

\bibitem[{Chan and Ghosh(2022)}]{chan2022contentcontextfactorizedrepresentationsautomated}
David~M. Chan and Shalini Ghosh. 2022.
\newblock \href {https://arxiv.org/abs/2205.09872} {Content-context factorized representations for automated speech recognition}.
\newblock \emph{Preprint}, arXiv:2205.09872.

\bibitem[{Cheng et~al.(2020)Cheng, Hao, Dai, Liu, Gan, and Carin}]{pmlr-v119-cheng20b}
Pengyu Cheng, Weituo Hao, Shuyang Dai, Jiachang Liu, Zhe Gan, and Lawrence Carin. 2020.
\newblock \href {https://proceedings.mlr.press/v119/cheng20b.html} {{CLUB}: A contrastive log-ratio upper bound of mutual information}.
\newblock In \emph{Proceedings of the 37th International Conference on Machine Learning}, volume 119 of \emph{Proceedings of Machine Learning Research}, pages 1779--1788. PMLR.

\bibitem[{Cheng et~al.(2023)Cheng, Dong, Yue, Ko, Wang, and Zou}]{cheng2023m}
Xuxin Cheng, Qianqian Dong, Fengpeng Yue, Tom Ko, Mingxuan Wang, and Yuexian Zou. 2023.
\newblock M 3 st: Mix at three levels for speech translation.
\newblock In \emph{ICASSP 2023-2023 IEEE International Conference on Acoustics, Speech and Signal Processing (ICASSP)}, pages 1--5. IEEE.

\bibitem[{Chou et~al.(2019)Chou, Yeh, and Lee}]{chou2019one}
Ju-chieh Chou, Cheng-chieh Yeh, and Hung-yi Lee. 2019.
\newblock One-shot voice conversion by separating speaker and content representations with instance normalization.
\newblock \emph{arXiv preprint arXiv:1904.05742}.

\bibitem[{Di~Gangi et~al.(2019)Di~Gangi, Cattoni, Bentivogli, Negri, and Turchi}]{di-gangi-etal-2019-must}
Mattia~A. Di~Gangi, Roldano Cattoni, Luisa Bentivogli, Matteo Negri, and Marco Turchi. 2019.
\newblock \href {https://doi.org/10.18653/v1/N19-1202} {{M}u{ST}-{C}: a {M}ultilingual {S}peech {T}ranslation {C}orpus}.
\newblock In \emph{Proceedings of the 2019 Conference of the North {A}merican Chapter of the Association for Computational Linguistics: Human Language Technologies, Volume 1 (Long and Short Papers)}, pages 2012--2017, Minneapolis, Minnesota. Association for Computational Linguistics.

\bibitem[{Dong et~al.(2021)Dong, Ye, Wang, Zhou, Xu, Xu, and Li}]{dong2021listen}
Qianqian Dong, Rong Ye, Mingxuan Wang, Hao Zhou, Shuang Xu, Bo~Xu, and Lei Li. 2021.
\newblock Listen, understand and translate: Triple supervision decouples end-to-end speech-to-text translation.
\newblock In \emph{Proc. of AAAI}, volume~35, pages 12749--12759.

\bibitem[{Du et~al.(2022)Du, Zhang, Wang, Chen, Xie, and Xu}]{du2022regularizing}
Yichao Du, Zhirui Zhang, Weizhi Wang, Boxing Chen, Jun Xie, and Tong Xu. 2022.
\newblock Regularizing end-to-end speech translation with triangular decomposition agreement.
\newblock In \emph{Proc. of AAAI}, volume~36, pages 10590--10598.

\bibitem[{Fang and Feng(2023)}]{fang-feng-2023-back}
Qingkai Fang and Yang Feng. 2023.
\newblock \href {https://doi.org/10.18653/v1/2023.acl-long.251} {Back translation for speech-to-text translation without transcripts}.
\newblock In \emph{Proceedings of the 61st Annual Meeting of the Association for Computational Linguistics (Volume 1: Long Papers)}, pages 4567--4587, Toronto, Canada. Association for Computational Linguistics.

\bibitem[{Fang et~al.(2022)Fang, Ye, Li, Feng, and Wang}]{fang2022stemm}
Qingkai Fang, Rong Ye, Lei Li, Yang Feng, and Mingxuan Wang. 2022.
\newblock Stemm: Self-learning with speech-text manifold mixup for speech translation.
\newblock In \emph{Proceedings of the 60th Annual Meeting of the Association for Computational Linguistics (Volume 1: Long Papers)}, pages 7050--7062.

\bibitem[{Han et~al.(2021)Han, Wang, Ji, and Li}]{han2021learning}
Chi Han, Mingxuan Wang, Heng Ji, and Lei Li. 2021.
\newblock Learning shared semantic space for speech-to-text translation.
\newblock In \emph{Findings of the Association for Computational Linguistics: ACL-IJCNLP 2021}, pages 2214--2225.

\bibitem[{Ho~Chan et~al.(2022)Ho~Chan, Qian, Zhang, and Hasegawa-Johnson}]{9747763}
Chak Ho~Chan, Kaizhi Qian, Yang Zhang, and Mark Hasegawa-Johnson. 2022.
\newblock \href {https://doi.org/10.1109/ICASSP43922.2022.9747763} {Speechsplit2.0: Unsupervised speech disentanglement for voice conversion without tuning autoencoder bottlenecks}.
\newblock In \emph{ICASSP 2022 - 2022 IEEE International Conference on Acoustics, Speech and Signal Processing (ICASSP)}, pages 6332--6336.

\bibitem[{Inaguma et~al.(2020)Inaguma, Kiyono, Duh, Karita, Yalta, Hayashi, and Watanabe}]{inaguma2020espnet}
Hirofumi Inaguma, Shun Kiyono, Kevin Duh, Shigeki Karita, Nelson Yalta, Tomoki Hayashi, and Shinji Watanabe. 2020.
\newblock Espnet-st: All-in-one speech translation toolkit.
\newblock In \emph{Proceedings of the 58th Annual Meeting of the Association for Computational Linguistics: System Demonstrations}, pages 302--311.

\bibitem[{Indurthi et~al.(2023)Indurthi, Chollampatt, Agrawal, and Turchi}]{indurthi-etal-2023-clad}
Sathish Indurthi, Shamil Chollampatt, Ravi Agrawal, and Marco Turchi. 2023.
\newblock \href {https://doi.org/10.18653/v1/2023.emnlp-main.560} {{CLAD}-{ST}: Contrastive learning with adversarial data for robust speech translation}.
\newblock In \emph{Proceedings of the 2023 Conference on Empirical Methods in Natural Language Processing}, pages 9049--9056, Singapore. Association for Computational Linguistics.

\bibitem[{Indurthi et~al.(2020)Indurthi, Han, Lakumarapu, Lee, Chung, Kim, and Kim}]{indurthi2020end}
Sathish Indurthi, Houjeung Han, Nikhil~Kumar Lakumarapu, Beomseok Lee, Insoo Chung, Sangha Kim, and Chanwoo Kim. 2020.
\newblock End-end speech-to-text translation with modality agnostic meta-learning.
\newblock In \emph{Proc. of ICASSP}, pages 7904--7908. IEEE.

\bibitem[{Kingma and Ba(2017)}]{kingma2017adam}
Diederik~P. Kingma and Jimmy Ba. 2017.
\newblock \href {https://arxiv.org/abs/1412.6980} {Adam: A method for stochastic optimization}.
\newblock \emph{Preprint}, arXiv:1412.6980.

\bibitem[{Kong et~al.(2023)Kong, Xu, and Mei}]{s23167282}
Yeqiu Kong, Zhongwei Xu, and Meng Mei. 2023.
\newblock \href {https://doi.org/10.3390/s23167282} {Cross-domain sentiment analysis based on feature projection and multi-source attention in iot}.
\newblock \emph{Sensors}, 23(16).

\bibitem[{Kudo and Richardson(2018)}]{kudo-richardson-2018-sentencepiece}
Taku Kudo and John Richardson. 2018.
\newblock \href {https://doi.org/10.18653/v1/D18-2012} {{S}entence{P}iece: A simple and language independent subword tokenizer and detokenizer for neural text processing}.
\newblock In \emph{Proceedings of the 2018 Conference on Empirical Methods in Natural Language Processing: System Demonstrations}, pages 66--71, Brussels, Belgium. Association for Computational Linguistics.

\bibitem[{Lam et~al.(2022)Lam, Schamoni, and Riezler}]{lam2022sample}
Tsz~Kin Lam, Shigehiko Schamoni, and Stefan Riezler. 2022.
\newblock Sample, translate, recombine: Leveraging audio alignments for data augmentation in end-to-end speech translation.
\newblock In \emph{Proceedings of the 60th Annual Meeting of the Association for Computational Linguistics (Volume 2: Short Papers)}, pages 245--254.

\bibitem[{Le et~al.(2023)Le, Gong, Wang, Pino, Lecouteux, and Schwab}]{pmlr-v202-le23a}
Phuong-Hang Le, Hongyu Gong, Changhan Wang, Juan Pino, Benjamin Lecouteux, and Didier Schwab. 2023.
\newblock \href {https://proceedings.mlr.press/v202/le23a.html} {Pre-training for speech translation: {CTC} meets optimal transport}.
\newblock In \emph{Proceedings of the 40th International Conference on Machine Learning}, volume 202 of \emph{Proceedings of Machine Learning Research}, pages 18667--18685. PMLR.

\bibitem[{Lee et~al.(2021)Lee, Park, and Kim}]{lee2021styler}
Keon Lee, Kyumin Park, and Daeyoung Kim. 2021.
\newblock \href {https://arxiv.org/abs/2103.09474} {Styler: Style factor modeling with rapidity and robustness via speech decomposition for expressive and controllable neural text to speech}.
\newblock \emph{Preprint}, arXiv:2103.09474.

\bibitem[{Lei et~al.(2023)Lei, Xue, Zhao, Sun, Zhu, Lin, and Xiong}]{lei-etal-2023-ckdst}
Yikun Lei, Zhengshan Xue, Xiaohu Zhao, Haoran Sun, Shaolin Zhu, Xiaodong Lin, and Deyi Xiong. 2023.
\newblock \href {https://doi.org/10.18653/v1/2023.findings-acl.195} {{CKDST}: Comprehensively and effectively distill knowledge from machine translation to end-to-end speech translation}.
\newblock In \emph{Findings of the Association for Computational Linguistics: ACL 2023}, pages 3123--3137, Toronto, Canada. Association for Computational Linguistics.

\bibitem[{Li et~al.(2023)Li, Wang, and Wu}]{li2023unified}
Wenbiao Li, Ziyang Wang, and Yunfang Wu. 2023.
\newblock \href {https://arxiv.org/abs/2210.10305} {A unified neural network model for readability assessment with feature projection and length-balanced loss}.
\newblock \emph{Preprint}, arXiv:2210.10305.

\bibitem[{Liu et~al.(2019)Liu, Xiong, Zhang, He, Wu, Wang, and Zong}]{liu2019end}
Yuchen Liu, Hao Xiong, Jiajun Zhang, Zhongjun He, Hua Wu, Haifeng Wang, and Chengqing Zong. 2019.
\newblock End-to-end speech translation with knowledge distillation.
\newblock \emph{Proc. Interspeech 2019}, pages 1128--1132.

\bibitem[{Liu et~al.(2020)Liu, Zhu, Zhang, and Zong}]{liu2020bridging}
Yuchen Liu, Junnan Zhu, Jiajun Zhang, and Chengqing Zong. 2020.
\newblock \href {https://arxiv.org/abs/2010.14920} {Bridging the modality gap for speech-to-text translation}.
\newblock \emph{Preprint}, arXiv:2010.14920.

\bibitem[{Mi et~al.(2022)Mi, Xie, and Zhang}]{mi2022improving}
Chenggang Mi, Lei Xie, and Yanning Zhang. 2022.
\newblock Improving data augmentation for low resource speech-to-text translation with diverse paraphrasing.
\newblock \emph{Neural Networks}, 148:194--205.

\bibitem[{Ott et~al.(2019)Ott, Edunov, Baevski, Fan, Gross, Ng, Grangier, and Auli}]{ott2019fairseq}
Myle Ott, Sergey Edunov, Alexei Baevski, Angela Fan, Sam Gross, Nathan Ng, David Grangier, and Michael Auli. 2019.
\newblock fairseq: A fast, extensible toolkit for sequence modeling.
\newblock In \emph{Proceedings of NAACL-HLT 2019: Demonstrations}.

\bibitem[{Panayotov et~al.(2015)Panayotov, Chen, Povey, and Khudanpur}]{librispeech}
Vassil Panayotov, Guoguo Chen, Daniel Povey, and Sanjeev Khudanpur. 2015.
\newblock \href {https://doi.org/10.1109/ICASSP.2015.7178964} {Librispeech: An asr corpus based on public domain audio books}.
\newblock In \emph{2015 IEEE International Conference on Acoustics, Speech and Signal Processing (ICASSP)}, pages 5206--5210.

\bibitem[{Park et~al.(2019)Park, Chan, Zhang, Chiu, Zoph, Cubuk, and Le}]{park19e_interspeech}
Daniel~S. Park, William Chan, Yu~Zhang, Chung-Cheng Chiu, Barret Zoph, Ekin~D. Cubuk, and Quoc~V. Le. 2019.
\newblock \href {https://doi.org/10.21437/Interspeech.2019-2680} {{SpecAugment: A Simple Data Augmentation Method for Automatic Speech Recognition}}.
\newblock In \emph{Proc. Interspeech 2019}, pages 2613--2617.

\bibitem[{Pino et~al.(2020)Pino, Xu, Ma, Dousti, and Tang}]{pino2020self}
Juan Pino, Qiantong Xu, Xutai Ma, Mohammad~Javad Dousti, and Yun Tang. 2020.
\newblock Self-training for end-to-end speech translation.

\bibitem[{Popovi{\'c}(2017)}]{popovic-2017-chrf}
Maja Popovi{\'c}. 2017.
\newblock \href {https://doi.org/10.18653/v1/W17-4770} {chr{F}++: words helping character n-grams}.
\newblock In \emph{Proceedings of the Second Conference on Machine Translation}, pages 612--618, Copenhagen, Denmark. Association for Computational Linguistics.

\bibitem[{Post(2018)}]{post-2018-call}
Matt Post. 2018.
\newblock \href {https://www.aclweb.org/anthology/W18-6319} {A call for clarity in reporting {BLEU} scores}.
\newblock In \emph{Proceedings of the Third Conference on Machine Translation: Research Papers}, pages 186--191, Belgium, Brussels. Association for Computational Linguistics.

\bibitem[{Qian et~al.(2020)Qian, Zhang, Chang, Hasegawa-Johnson, and Cox}]{pmlr-v119-qian20a}
Kaizhi Qian, Yang Zhang, Shiyu Chang, Mark Hasegawa-Johnson, and David Cox. 2020.
\newblock \href {https://proceedings.mlr.press/v119/qian20a.html} {Unsupervised speech decomposition via triple information bottleneck}.
\newblock In \emph{Proceedings of the 37th International Conference on Machine Learning}, volume 119 of \emph{Proceedings of Machine Learning Research}, pages 7836--7846. PMLR.

\bibitem[{Qin et~al.(2020)Qin, Hu, and Liu}]{qin-etal-2020-feature}
Qi~Qin, Wenpeng Hu, and Bing Liu. 2020.
\newblock \href {https://doi.org/10.18653/v1/2020.acl-main.726} {Feature projection for improved text classification}.
\newblock In \emph{Proceedings of the 58th Annual Meeting of the Association for Computational Linguistics}, pages 8161--8171, Online. Association for Computational Linguistics.

\bibitem[{Rei et~al.(2022)Rei, C.~de Souza, Alves, Zerva, Farinha, Glushkova, Lavie, Coheur, and Martins}]{rei-etal-2022-comet}
Ricardo Rei, Jos{\'e}~G. C.~de Souza, Duarte Alves, Chrysoula Zerva, Ana~C Farinha, Taisiya Glushkova, Alon Lavie, Luisa Coheur, and Andr{\'e} F.~T. Martins. 2022.
\newblock \href {https://aclanthology.org/2022.wmt-1.52} {{COMET}-22: Unbabel-{IST} 2022 submission for the metrics shared task}.
\newblock In \emph{Proceedings of the Seventh Conference on Machine Translation (WMT)}, pages 578--585, Abu Dhabi, United Arab Emirates (Hybrid). Association for Computational Linguistics.

\bibitem[{Skerry-Ryan et~al.(2018)Skerry-Ryan, Battenberg, Xiao, Wang, Stanton, Shor, Weiss, Clark, and Saurous}]{pmlr-v80-skerry-ryan18a}
RJ~Skerry-Ryan, Eric Battenberg, Ying Xiao, Yuxuan Wang, Daisy Stanton, Joel Shor, Ron Weiss, Rob Clark, and Rif~A. Saurous. 2018.
\newblock \href {https://proceedings.mlr.press/v80/skerry-ryan18a.html} {Towards end-to-end prosody transfer for expressive speech synthesis with tacotron}.
\newblock In \emph{Proceedings of the 35th International Conference on Machine Learning}, volume~80 of \emph{Proceedings of Machine Learning Research}, pages 4693--4702. PMLR.

\bibitem[{Sperber et~al.(2017)Sperber, Neubig, Niehues, and Waibel}]{sperber2017neural}
Matthias Sperber, Graham Neubig, Jan Niehues, and Alex Waibel. 2017.
\newblock Neural lattice-to-sequence models for uncertain inputs.
\newblock In \emph{Proc. of EMNLP}.

\bibitem[{Sperber et~al.(2019)Sperber, Neubig, Pham, and Waibel}]{sperber-etal-2019-self}
Matthias Sperber, Graham Neubig, Ngoc-Quan Pham, and Alex Waibel. 2019.
\newblock \href {https://doi.org/10.18653/v1/P19-1115} {Self-attentional models for lattice inputs}.
\newblock In \emph{Proceedings of the 57th Annual Meeting of the Association for Computational Linguistics}, pages 1185--1197, Florence, Italy. Association for Computational Linguistics.

\bibitem[{Tang et~al.(2021)Tang, Pino, Li, Wang, and Genzel}]{tang2021improving}
Yun Tang, Juan Pino, Xian Li, Changhan Wang, and Dmitriy Genzel. 2021.
\newblock Improving speech translation by understanding and learning from the auxiliary text translation task.
\newblock In \emph{Proceedings of the 59th Annual Meeting of the Association for Computational Linguistics and the 11th International Joint Conference on Natural Language Processing (Volume 1: Long Papers)}, pages 4252--4261.

\bibitem[{Van~der Maaten and Hinton(2008)}]{van2008visualizing}
Laurens Van~der Maaten and Geoffrey Hinton. 2008.
\newblock Visualizing data using t-sne.
\newblock \emph{Journal of machine learning research}, 9(11).

\bibitem[{Vydana et~al.(2021)Vydana, Karafi{\'a}t, Zmolikova, Burget, and {\v{C}}ernock{\`y}}]{vydana2021jointly}
Hari~Krishna Vydana, Martin Karafi{\'a}t, Katerina Zmolikova, Luk{\'a}{\v{s}} Burget, and Honza {\v{C}}ernock{\`y}. 2021.
\newblock Jointly trained transformers models for spoken language translation.
\newblock In \emph{ICASSP 2021-2021 IEEE International Conference on Acoustics, Speech and Signal Processing (ICASSP)}, pages 7513--7517. IEEE.

\bibitem[{Wang et~al.(2020{\natexlab{a}})Wang, Tang, Ma, Wu, Okhonko, and Pino}]{wang2020fairseq}
Changhan Wang, Yun Tang, Xutai Ma, Anne Wu, Dmytro Okhonko, and Juan Pino. 2020{\natexlab{a}}.
\newblock Fairseq s2t: Fast speech-to-text modeling with fairseq.
\newblock In \emph{Proceedings of the 1st Conference of the Asia-Pacific Chapter of the Association for Computational Linguistics and the 10th International Joint Conference on Natural Language Processing: System Demonstrations}, pages 33--39.

\bibitem[{Wang et~al.(2020{\natexlab{b}})Wang, Wu, and Pino}]{wang2020covost}
Changhan Wang, Anne Wu, and Juan Pino. 2020{\natexlab{b}}.
\newblock \href {https://arxiv.org/abs/2007.10310} {Covost 2: A massively multilingual speech-to-text translation corpus}.
\newblock \emph{Preprint}, arXiv:2007.10310.

\bibitem[{Wang et~al.(2020{\natexlab{c}})Wang, Wu, Liu, Yang, and Zhou}]{Wang_Wu_Liu_Yang_Zhou_2020}
Chengyi Wang, Yu~Wu, Shujie Liu, Zhenglu Yang, and Ming Zhou. 2020{\natexlab{c}}.
\newblock \href {https://doi.org/10.1609/aaai.v34i05.6452} {Bridging the gap between pre-training and fine-tuning for end-to-end speech translation}.
\newblock \emph{Proceedings of the AAAI Conference on Artificial Intelligence}, 34(05):9161--9168.

\bibitem[{Xia et~al.(2023)Xia, Huang, Zhu, and Zhao}]{NEURIPS2023_c89f0984}
Yan Xia, Hai Huang, Jieming Zhu, and Zhou Zhao. 2023.
\newblock \href {https://proceedings.neurips.cc/paper_files/paper/2023/file/c89f09849eb5af489abb122394ff0f0b-Paper-Conference.pdf} {Achieving cross modal generalization with multimodal unified representation}.
\newblock In \emph{Advances in Neural Information Processing Systems}, volume~36, pages 63529--63541. Curran Associates, Inc.

\bibitem[{Xie et~al.(2022)Xie, Pun, and Lam}]{9682736}
Jiu-Cheng Xie, Chi-Man Pun, and Kin-Man Lam. 2022.
\newblock \href {https://doi.org/10.1109/TIFS.2022.3142998} {Implicit and explicit feature purification for age-invariant facial representation learning}.
\newblock \emph{IEEE Transactions on Information Forensics and Security}, 17:399--412.

\bibitem[{Xiong et~al.(2020)Xiong, Yang, He, Zheng, Zheng, Xing, Zhang, Lan, Wang, and Liu}]{pmlr-v119-xiong20b}
Ruibin Xiong, Yunchang Yang, Di~He, Kai Zheng, Shuxin Zheng, Chen Xing, Huishuai Zhang, Yanyan Lan, Liwei Wang, and Tieyan Liu. 2020.
\newblock \href {https://proceedings.mlr.press/v119/xiong20b.html} {On layer normalization in the transformer architecture}.
\newblock In \emph{Proceedings of the 37th International Conference on Machine Learning}, volume 119 of \emph{Proceedings of Machine Learning Research}, pages 10524--10533. PMLR.

\bibitem[{Xu et~al.(2021)Xu, Hu, Li, Zhang, Huang, Ju, Xiao, and Zhu}]{xu2021stacked}
Chen Xu, Bojie Hu, Yanyang Li, Yuhao Zhang, Shen Huang, Qi~Ju, Tong Xiao, and Jingbo Zhu. 2021.
\newblock Stacked acoustic-and-textual encoding: Integrating the pre-trained models into speech translation encoders.
\newblock In \emph{Proc. ACL}, pages 2619--2630.

\bibitem[{Yan et~al.(2024)Yan, Chang, Anastasopoulos, Fujita, and Watanabe}]{10447926}
Brian Yan, Xuankai Chang, Antonios Anastasopoulos, Yuya Fujita, and Shinji Watanabe. 2024.
\newblock \href {https://doi.org/10.1109/ICASSP48485.2024.10447926} {Cross-modal multi-tasking for speech-to-text translation via hard parameter sharing}.
\newblock In \emph{ICASSP 2024 - 2024 IEEE International Conference on Acoustics, Speech and Signal Processing (ICASSP)}, pages 11941--11945.

\bibitem[{Yang et~al.(2022)Yang, Tantrawenith, Zhuang, Wu, Sun, Wang, Cheng, Tang, Zhao, Wang, and Meng}]{yang22f_interspeech}
SiCheng Yang, Methawee Tantrawenith, Haolin Zhuang, Zhiyong Wu, Aolan Sun, Jianzong Wang, Ning Cheng, Huaizhen Tang, Xintao Zhao, Jie Wang, and Helen Meng. 2022.
\newblock \href {https://doi.org/10.21437/Interspeech.2022-571} {{Speech Representation Disentanglement with Adversarial Mutual Information Learning for One-shot Voice Conversion}}.
\newblock In \emph{Proc. Interspeech 2022}, pages 2553--2557.

\bibitem[{Yang et~al.(2021)Yang, Hira, Ni, Chourdia, Astafurov, Chen, Yeh, Puhrsch, Pollack, Genzel, Greenberg, Yang, Lian, Mahadeokar, Hwang, Chen, Goldsborough, Roy, Narenthiran, Watanabe, Chintala, Quenneville-Bélair, and Shi}]{yang2021torchaudio}
Yao-Yuan Yang, Moto Hira, Zhaoheng Ni, Anjali Chourdia, Artyom Astafurov, Caroline Chen, Ching-Feng Yeh, Christian Puhrsch, David Pollack, Dmitriy Genzel, Donny Greenberg, Edward~Z. Yang, Jason Lian, Jay Mahadeokar, Jeff Hwang, Ji~Chen, Peter Goldsborough, Prabhat Roy, Sean Narenthiran, Shinji Watanabe, Soumith Chintala, Vincent Quenneville-Bélair, and Yangyang Shi. 2021.
\newblock Torchaudio: Building blocks for audio and speech processing.
\newblock \emph{arXiv preprint arXiv:2110.15018}.

\bibitem[{Ye et~al.(2021{\natexlab{a}})Ye, Wang, and Li}]{ye21_interspeech}
Rong Ye, Mingxuan Wang, and Lei Li. 2021{\natexlab{a}}.
\newblock \href {https://doi.org/10.21437/Interspeech.2021-1065} {{End-to-End Speech Translation via Cross-Modal Progressive Training}}.
\newblock In \emph{Proc. Interspeech 2021}, pages 2267--2271.

\bibitem[{Ye et~al.(2021{\natexlab{b}})Ye, Wang, and Li}]{ye2021end}
Rong Ye, Mingxuan Wang, and Lei Li. 2021{\natexlab{b}}.
\newblock End-to-end speech translation via cross-modal progressive training.
\newblock \emph{arXiv preprint arXiv:2104.10380}.

\bibitem[{Ye et~al.(2022)Ye, Wang, and Li}]{ye2022cross}
Rong Ye, Mingxuan Wang, and Lei Li. 2022.
\newblock Cross-modal contrastive learning for speech translation.
\newblock In \emph{Proceedings of the 2022 Conference of the North American Chapter of the Association for Computational Linguistics: Human Language Technologies}, pages 5099--5113.

\bibitem[{Yuan et~al.(2024)Yuan, Zhou, and Shi}]{yuan2024memory}
Yuxuan Yuan, Yue Zhou, and Xiaodong Shi. 2024.
\newblock Memory-augmented speech-to-text translation with multi-scale context translation strategy.
\newblock In \emph{ICASSP 2024-2024 IEEE International Conference on Acoustics, Speech and Signal Processing (ICASSP)}, pages 12727--12731. IEEE.

\bibitem[{Zhang et~al.(2022)Zhang, Haddow, and Sennrich}]{pmlr-v162-zhang22i}
Biao Zhang, Barry Haddow, and Rico Sennrich. 2022.
\newblock \href {https://proceedings.mlr.press/v162/zhang22i.html} {Revisiting end-to-end speech-to-text translation from scratch}.
\newblock In \emph{Proceedings of the 39th International Conference on Machine Learning}, volume 162 of \emph{Proceedings of Machine Learning Research}, pages 26193--26205. PMLR.

\bibitem[{Zhang et~al.(2023{\natexlab{a}})Zhang, Ye, Ko, Wang, and Zhou}]{zhang2023dub}
Dong Zhang, Rong Ye, Tom Ko, Mingxuan Wang, and Yaqian Zhou. 2023{\natexlab{a}}.
\newblock Dub: Discrete unit back-translation for speech translation.
\newblock In \emph{Findings of the Association for Computational Linguistics: ACL 2023}, pages 7147--7164.

\bibitem[{Zhang et~al.(2023{\natexlab{b}})Zhang, Si, Chen, Zhang, Yang, Qu, and Li}]{10096899}
Hao Zhang, Nianwen Si, Yaqi Chen, Wenlin Zhang, Xukui Yang, Dan Qu, and Zhen Li. 2023{\natexlab{b}}.
\newblock \href {https://doi.org/10.1109/ICASSP49357.2023.10096899} {Decoupled non-parametric knowledge distillation for end-to-end speech translation}.
\newblock In \emph{ICASSP 2023 - 2023 IEEE International Conference on Acoustics, Speech and Signal Processing (ICASSP)}, pages 1--5.

\bibitem[{Zhang et~al.(2024)Zhang, Kou, Li, Xu, Zhang, Xiao, and Zhu}]{10447494}
Yuhao Zhang, Kaiqi Kou, Bei Li, Chen Xu, Chunliang Zhang, Tong Xiao, and Jingbo Zhu. 2024.
\newblock \href {https://doi.org/10.1109/ICASSP48485.2024.10447494} {Soft alignment of modality space for end-to-end speech translation}.
\newblock In \emph{ICASSP 2024 - 2024 IEEE International Conference on Acoustics, Speech and Signal Processing (ICASSP)}, pages 11041--11045.

\bibitem[{Zhang et~al.(2023{\natexlab{c}})Zhang, Xu, Li, Chen, Xiao, Zhang, and Zhu}]{zhang2023rethinking}
Yuhao Zhang, Chen Xu, Bei Li, Hao Chen, Tong Xiao, Chunliang Zhang, and Jingbo Zhu. 2023{\natexlab{c}}.
\newblock Rethinking and improving multi-task learning for end-to-end speech translation.
\newblock In \emph{Proceedings of the 2023 Conference on Empirical Methods in Natural Language Processing}, pages 10753--10765.

\bibitem[{Zhao et~al.(2023)Zhao, Sun, Lei, Zhu, and Xiong}]{zhao-etal-2023-ccsrd}
Xiaohu Zhao, Haoran Sun, Yikun Lei, Shaolin Zhu, and Deyi Xiong. 2023.
\newblock \href {https://doi.org/10.18653/v1/2023.findings-emnlp.394} {{CCSRD}: Content-centric speech representation disentanglement learning for end-to-end speech translation}.
\newblock In \emph{Findings of the Association for Computational Linguistics: EMNLP 2023}, pages 5920--5932, Singapore. Association for Computational Linguistics.

\bibitem[{Zhou et~al.(2023)Zhou, Fang, and Feng}]{zhou2023cmot}
Yan Zhou, Qingkai Fang, and Yang Feng. 2023.
\newblock Cmot: Cross-modal mixup via optimal transport for speech translation.
\newblock In \emph{Proceedings of the 61st Annual Meeting of the Association for Computational Linguistics (Volume 1: Long Papers)}, pages 7873--7887.

\bibitem[{Zhou et~al.(2024)Zhou, Yuan, and Shi}]{zhou2024multitask}
Yue Zhou, Yuxuan Yuan, and Xiaodong Shi. 2024.
\newblock A multitask co-training framework for improving speech translation by leveraging speech recognition and machine translation tasks.
\newblock \emph{Neural Computing and Applications}, pages 1--16.

\bibitem[{Zhu et~al.(2023)Zhu, Zhang, Lin, Sun, and Cheng}]{10008072}
Ziyue Zhu, Zhao Zhang, Zheng Lin, Xing Sun, and Ming-Ming Cheng. 2023.
\newblock \href {https://doi.org/10.1109/TPAMI.2023.3234586} {Co-salient object detection with co-representation purification}.
\newblock \emph{IEEE Transactions on Pattern Analysis and Machine Intelligence}, 45(7):8193--8205.

\end{thebibliography}

\newpage
\appendix
\section{Preliminary Experiment Details}
\label{appendix:preliminary}
Our preliminary experiment are implemented base on XSTNet~\citep{ye2021end}, we first train this model from scratch with MuST-C~\cite{di-gangi-etal-2019-must} En-De training set. Then we convert the MuST-C training samples with a one-shot voice conversion system~\citep{chou2019one}\footnote{\url{https://github.com/jjery2243542/adaptive_voice_conversion}}. Specifically, we randomly collect 1000 samples from Common Voice~\citep{commonvoice:2020} dataset with different speakers. For each sample in MuST-C En-De, we randomly select 1 sample from 1000 CommonVoice samples to perform on-shot voice conversion.

\section{Statistics of all datasets}
\label{appendix:datasets}
\begin{table}[htb]
    \centering
    % \small
    \resizebox{\columnwidth}{!}{
    \begin{tabular}{c|cc|cc}
        \toprule
         & \multicolumn{2}{c|}{\textbf{ST (MuST-C)}} & \multicolumn{2}{c}{\textbf{MT}}  \\
         \textbf{Lang} & hours & \#sents & name & \#sents \\
         \midrule
         \multicolumn{5}{c}{\textbf{MuST-C En $\to$ X}} \\
         \midrule
         \textbf{En-De} & 408 & 234K & WMT16 & 4.6M \\
         \textbf{En-Ru} & 489 & 270K & WMT14 & 40.8M  \\
         \textbf{En-Es} & 504 & 270K & WMT13 & 15.2M \\
         \textbf{En-It} & 465 & 258K & - & - \\
         \textbf{En-Fr} & 492 & 280K & - & - \\
         \textbf{En-Ro} & 432 & 240K & - & - \\
         \textbf{En-Pt} & 385 & 211K & - & - \\
         \textbf{En-Nl} & 442 & 253K & - & - \\
         \midrule
         \multicolumn{5}{c}{\textbf{CoVoST-2 X $\to$ En}} \\
         \midrule
         \textbf{De-En} & 184 & - & - & - \\
         \textbf{Fr-En} & 264 & - & - & - \\
        \bottomrule
    \end{tabular}}
    \caption{Statistics of all the datasets we used.}
    \label{tab:data}
\end{table}
\section{Experimental Details}
\label{appendix:Experiment}

\noindent\textbf{Training and Implementation Details}
For speech input, we use the raw 16-bit 16kHz mono-channel audio waveform. Training set samples with speech frames greater than 480,000 or less than 1,000 are removed. For each translation direction, we employ the unigram sentencepiece~\citep{kudo-richardson-2018-sentencepiece} model to build a subword vocabulary with a size of 10000 on the text data from the training set, the dictionary is shared across source and target languages.

We set the hidden size to 512, the FFN hidden dimension to 2048, and 8 attention heads. We set the number of layers of CA-Enc, CI-Enc, T-Enc, and Decoder to 1,1,5, and 6, respectively.
Both the classifiers in the OPP module employ the same architecture that consists of two linear layers with ReLU activation and a softmax classification layer, the inner hidden size of the linear layer is set to 1024.
According to the settings demonstrated above, our model has approximately 165 million trainable parameters.
We use the Adam~\citep{kingma2017adam} optimizer and inverse square root learning schedule with 4k warm-up updates.

We set the learning rate to 1e-4, dropout to $0.1$, and label smoothing value to $0.1$. 
We save a checkpoint at the end of each epoch and the training will early stop if the BLEU scores don't increase for 10 epochs on the dev set.
During inference, we average the model parameters on the last 10 checkpoints based on the performance on the dev set and adopt the beam search strategy with beam size 10.
The length penalty is set to $1.0$, $1.0$, $0.5$, $0.2$, $0.3$, $0.5$, $1.0$, and $1.2$ for En to De, Fr, Ru, Es, It, Nl, Pt, and Ro, respectively.
To perform a fair comparison with other models, we calculate and report case-sensitive detokenized BLEU scores using sacreBLEU\footnote{\url{https://github.com/mjpost/sacrebleu}, BLEU Signature: nrefs:1 | bs:1000 | seed:12345 | case:mixed | eff:no | tok:13a | smooth:exp | version:2.0.0}~\citep{post-2018-call} on tst-COMMON set.
We also provide the ChrF++\footnote{ChrF2++ Signature: nrefs:1 | bs:1000 | seed:12345 | case:mixed | eff:yes | nc:6 | nw:2 | space:no | version:2.0.0}, and COMET~\citep{rei-etal-2022-comet} scores with \textit{wmt22-comet-da} model.
We train all models with 2 Nvidia A40 GPUs, the training takes about 2 days to converge.

\noindent\textbf{Baselines} We compared our approach with several strong end-to-end ST systems under multitask setting including: our baseline model XSTnet~\cite{ye21_interspeech}, Memory-ST~\cite{yuan2024memory},  STEMM~\cite{fang2022stemm}, ConST~\cite{ye2022cross}, MCTN~\cite{zhou2024multitask}, Siamese-PT~\citep{pmlr-v202-le23a}, CCSRD~\cite{zhao-etal-2023-ccsrd}, M$^{3}$ST~\cite{cheng2023m}, and CMOT~\cite{zhou2023cmot}.
We also compared our method to other methods without the use of transcription data, including Transformer-ST with ASR pre-training~\cite{wang2020covost}, Revisit ST~\cite{pino2020self}, W2V2-Transformer~\cite{fang2022stemm}, and CCSRD~\cite{zhao-etal-2023-ccsrd}, DUB~\citep{zhang2023dub}, and BT4ST~\citep{fang-feng-2023-back}.
Note that although we compare our model with DUB and BT4ST under \textit{transcript-free} setting, these two models utilized additional MT training data to generate source speech or discrete audio tokens.
In addition, we compared our approach to these baseline systems that use additional MT data.

\label{sec:appendix}

\section{Variational Mutual Information Upper-bound Estimation}
\label{appendix:vclub}
\begin{algorithm}[h]
\caption{Mutual Information Upper-bound Minimization with vCLUB}
\begin{algorithmic}
    \REQUIRE{Content-agnostic representations $\mathbf{H}^{\beta^*}$; \\ Purified representation $\mathbf{H}^{\gamma}$; \\ Our model $\theta^m$; \\ Approximation network $\theta$; \\ $\mathcal{L}(\theta)=\frac{1}{N}\sum_N \log q_{\theta}(\mathbf{H}^{\gamma} \mid \mathbf{H}^{\beta^*})$;}
    \FOR {each training iteration}
    \WHILE{$\theta$ is not converge}
        \STATE update $\theta$ by maximizing $\mathcal{L}(\theta)$
    \ENDWHILE
    \STATE Estimate MI upper bound by Equation~\ref{eq:mi}
    \STATE Calculate the total loss (Equation~\ref{eq:mt})
    \STATE update $\theta^m$
    \ENDFOR
\end{algorithmic}
\end{algorithm}
To estimate the mutual information upper-bound between $\mathbf{H}^{\gamma}$ and $\mathbf{H}^{\beta^*}$, the contrastive log-ratio upper bound (CLUB)~\citep{pmlr-v119-cheng20b} is defined as:
\begin{equation}
\begin{split}
   \mathcal{I}_{\rm CLUB} = \mathbb{E}_{p(\mathbf{H}^{\gamma}, \mathbf{H}^{\beta^*})}[\log p(\mathbf{H}^{\gamma} \mid \mathbf{H}^{\beta^*})] \\ -\mathbb{E}_{p(\mathbf{H}^{\gamma})}\mathbb{E}_{p(\mathbf{H}^{\beta^*})}[\log p(\mathbf{H}^{\gamma} \mid \mathbf{H}^{\beta^*})],
\end{split}
\end{equation}
However, the conditional distribution $p(\mathbf{H}^{\gamma} \mid \mathbf{H}^{\beta^*})$ is unknown in our task. The CLUB was extended to more tasks by using a variational distribution $q_{\theta}(\mathbf{H}^{\gamma} \mid \mathbf{H}^{\beta^*})$ with an approximation network $\theta$. This variational CLUB (vCLUB) is consequently defined as:
\begin{equation}
\begin{split}
   \mathcal{I}_{\rm vCLUB} = \mathbb{E}_{p(\mathbf{H}^{\gamma}, \mathbf{H}^{\beta^*})}[\log q_{\theta}(\mathbf{H}^{\gamma} \mid \mathbf{H}^{\beta^*})] \\ -\mathbb{E}_{p(\mathbf{H}^{\gamma})}\mathbb{E}_{p(\mathbf{H}^{\beta^*})}[\log q_{\theta}(\mathbf{H}^{\gamma} \mid \mathbf{H}^{\beta^*})],
\end{split}
\end{equation}
In practice, the approximation network $\theta$ consists of two sub-networks (both are stacks of 5 linear layers with activation functions) that were used to model the posterior $p(\mathbf{H}^{\gamma} \mid \mathbf{H}^{\beta^*})$ by predicting a set of means and variances. This approximation network possesses an independent optimizer and is optimized alternatively during training. Additionally, following~\citep{NEURIPS2023_c89f0984}, we tailor it to suit our task, the mutual information loss is defined as in Equation~\ref{eq:mi}.

The detailed optimization process is demonstrated in Algorithm 1, for every step: (1) speech features firstly forward in our main network to get the content-agnostic and content-relevant representations. (2) Then the approximation network was optimized by 10 steps to converge. Then it will estimate the mutual information and we can calculate the $L_{MI}$ with Equation~\ref{eq:mi}. (3) Our main network can perform backward and be optimized.

\section{Hyper-parameter Selection Experiments}
\label{appendix:hyper-parameter}
\begin{figure}[h]
  \includegraphics[width=\columnwidth]{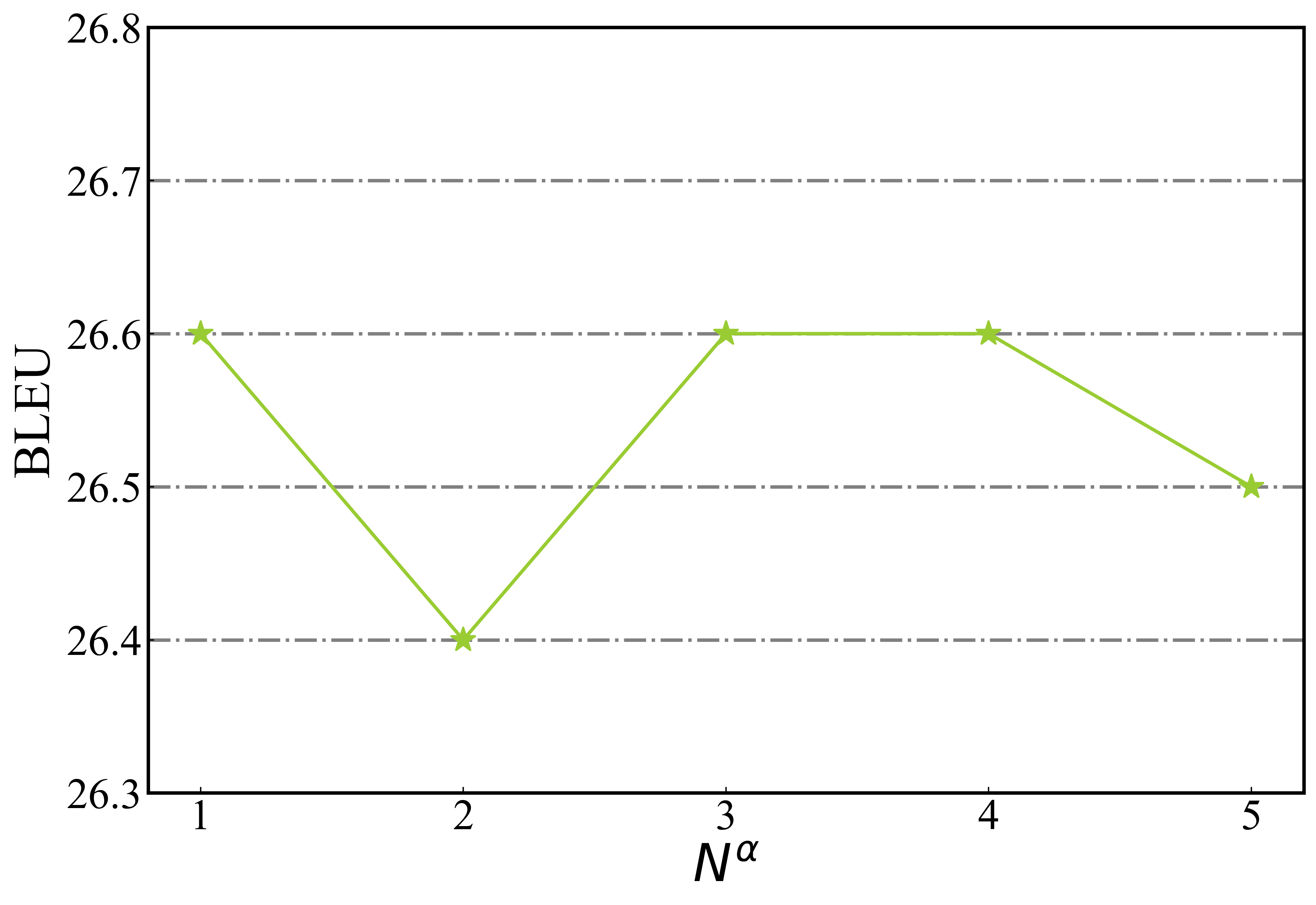}
  \caption{BLEU scores with different number of CA-Enc layers $N^{\alpha}$ on MuST-C En-De tst-COMMON set. Here the x-axis is the number of layers.}
  \label{fig:alpha.png}
\end{figure}
\begin{figure}[h]
  \includegraphics[width=\columnwidth]{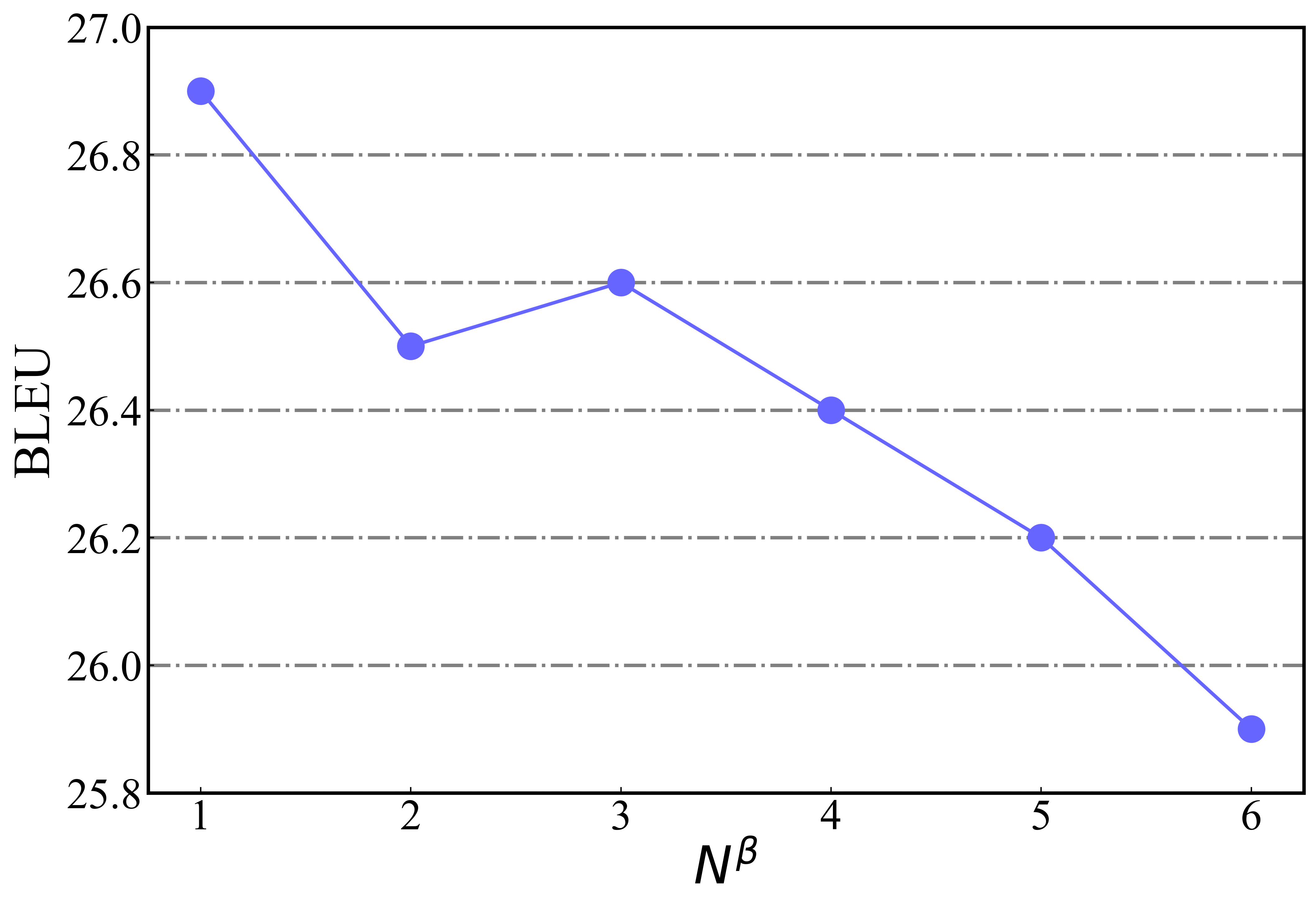}
  \caption{BLEU scores with different number of CI-Enc layers $N^{\beta}$ on MuST-C En-De tst-COMMON set. Here the x-axis is the number of layers.}
  \label{fig:beta.png}
\end{figure}
\begin{figure}[h]
  \includegraphics[width=\columnwidth]{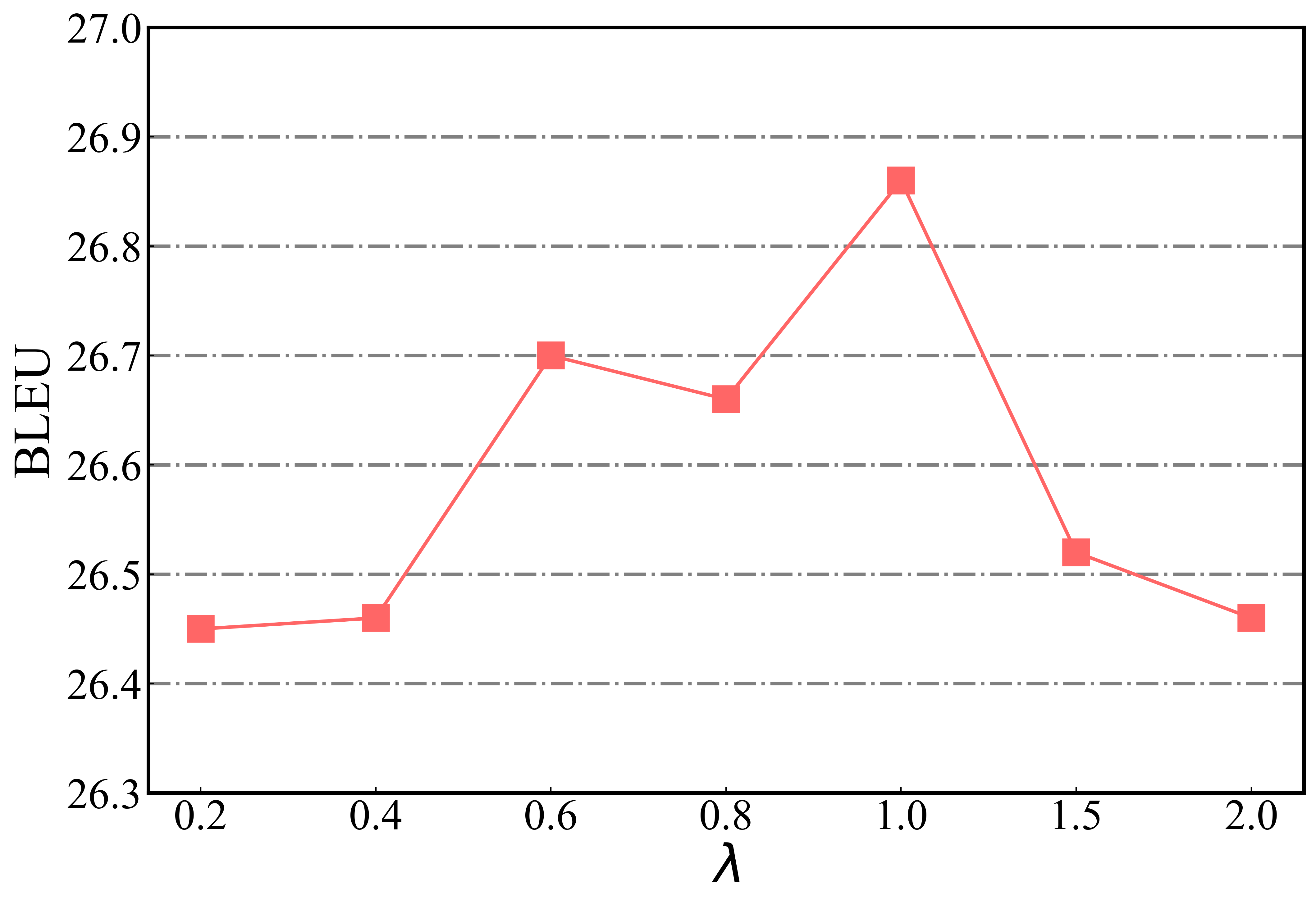}
  \caption{BLEU scores with different $\lambda_1$ on MuST-C En-De tst-COMMON set. Here the x-axis is the weight of $\mathcal{L}_{\rm CONSIS}$.}
  \label{fig:lambda.png}
\end{figure}
As demonstrated in Section~\ref{sec:exp_details}, we set the hyper-parameters $\lambda_1$, $\lambda_2$, $N^{\alpha}$, and $N^{\beta}$ to $1.0$, $0.01$, $1$, and $1$, respectively. We detail the hyper-parameter selection experiments in this section. Note that we set the number of T-Enc layers to $6-N^{\beta}$ to maintain an approximate model size with previous works for a fair comparison. Firstly, our initial setup of $\lambda_1$, $\lambda_2$, $N^{\alpha}$, and $N^{\beta}$ is $1.0$, $0.01$, $3$, and $3$, respectively. 

The results of selecting $N^{\alpha}$ are shown in Figure~\ref{fig:alpha.png}.
We find the performance has almost no changes as the number of layers increases, considering the computational expanse, we fix $N^{\alpha}$ to 1.
The results of selecting $N^{\beta}$ are shown in Figure~\ref{fig:beta.png}, for better translation quality, we set $N^{\beta}$ to 1.
We also demonstrate the selection experiments of $\lambda_1$ in Table~\ref{fig:lambda.png}, and we finally fix the $\lambda_1$ to 1.0. We didn't conduct experiments for selecting $\lambda_2$, following~\citet{yang22f_interspeech}, we set $\lambda_2$ to $0.01$, which means the performance of our model can still be improved by conducting more experiments to select $\lambda_2$.

\end{document}